\documentclass{article}
\usepackage[preprint]{neurips_2026}

\usepackage[utf8]{inputenc} % allow utf-8 input
\usepackage[T1]{fontenc}    % use 8-bit T1 fonts
\usepackage{hyperref}       % hyperlinks
\usepackage{url}            % simple URL typesetting
\usepackage{booktabs}       % professional-quality tables
\usepackage{amsfonts}       % blackboard math symbols
\usepackage{nicefrac}       % compact symbols for 1/2, etc.
\usepackage{microtype}      % microtypography
\usepackage{xcolor}         % colors

\hypersetup{
    colorlinks=true,
    linkcolor=blue,        % internal links (sections, refs)
    citecolor=teal,        % citations
    urlcolor=magenta,      % URLs
}

\usepackage{enumitem}
\usepackage{color}
\usepackage{soul}
\usepackage{wrapfig}
\usepackage{longtable}
\usepackage{caption}
\usepackage{multirow}
\usepackage{epsfig}
\usepackage{marvosym}
\usepackage{threeparttable}
\usepackage{setspace}
\usepackage{amsthm}
\usepackage{dutchcal}
\usepackage{makecell}
\usepackage[table]{xcolor}
\usepackage{amsmath}
\usepackage{cleveref}
\usepackage{float}
\usepackage[normalem]{ulem}
\usepackage{doi}
\usepackage{array}

\usepackage{colortbl}
\definecolor{ftblue}{RGB}{235,245,255} % very light blue

\newcommand{\best}[1]{\textbf{#1}}
\newcommand{\second}[1]{\underline{#1}}
\newcommand{\ftcell}[1]{\cellcolor{ftblue}#1}

\usepackage[most]{tcolorbox}
\definecolor{mygray}{RGB}{88,88,88}

\usepackage{graphicx}
\usepackage{subcaption}

\usepackage[toc,page,header]{appendix}
\usepackage{minitoc} 
\doparttoc % Tell to minitoc to generate a toc for the parts
\faketableofcontents % Run a fake tableofcontents command for the partocs
\title{How Do Electrocardiogram Models Scale?}

\author{
  Jiawei Li \\
  Uppsala University\\
  jiawei.li@it.uu.se
  \And
  Fabio Bonassi \\
  Uppsala University\\
  fabio.bonassi@it.uu.se
  \And
  Ming Jin \\
  Griffith University\\
  mingjinedu@gmail.com
  \And
  Stefan Gustafsson \\
  Uppsala University\\
  stefan.gustafsson@medsci.uu.se
  \And
  Johan Sundström \\
  Uppsala University\\
  johan.sundstrom@uu.se
  \And
  Thomas B. Schön \\
  Uppsala University \\
  thomas.schon@uu.se
  \And
  Antônio H. Ribeiro \\
  Uppsala University \\
  antonio.horta.ribeiro@it.uu.se
}

\begin{document}

\maketitle

\begin{abstract}
While scaling laws have established a fundamental framework for foundation models in natural language processing, their applicability to  electrocardiogram  (ECG) models remains poorly characterized. 
Indeed, recent studies do not always yield consistent downstream gains as one increases the model size or pre-training dataset size of ECG models,
leaving the exact roles of architectural inductive biases, pre-training paradigms, and expected improvements with size  largely unanswered.
In this work, we systematically investigate neural and loss-to-loss scaling laws within the ECG domain. 
By pre-training over $120$ models (ranging from $20$K to $200$M parameters) on the large-scale CODE dataset ($2.3$M records), we decouple the effects of model architecture (ResNet vs. Transformer) and pre-training paradigm, namely supervised learning (SL) versus self-supervised learning (SSL). 
We found that (\emph{i}) SL models are data-bottlenecked in-distribution, whereas SSL models scale robustly across both model and data sizes; (\emph{ii}) for out-of-distribution (OOD) generalization, ResNets are $1.3$ to $2.5$ times more parameter-efficient than Transformers, while SSL is up to $16$ times more data-efficient and achieves up to $7.6$ times higher transfer efficiency than SL on unseen clinical tasks; (\emph{iii}) across the observed scales, ResNet-based models generally achieve the lowest OOD loss, with SSL dominating on unseen clinical tasks and self-supervised Transformers overtaking at very large model sizes.
Our results suggest that the path to effective ECG foundation models lies in the strategic alignment of architecture and paradigm rather than brute-force scaling.
\end{abstract}

\section{Introduction }
Cardiac diseases are the main cause of mortality and morbidity worldwide~\citep{james2018global}, and the electrocardiogram (ECG) is a major diagnostic tool. It is low-cost, widely available, and used from basic care to emergency departments. 
ECG foundation models are deep learning architectures pre-trained on large-scale ECG datasets, typically via self-supervised or supervised learning paradigms~\citep{mckeen2024ecg, li2024electrocardiogram}.
This approach has been widely successful in various fields, and for ECGs it has the potential to enable the development of screening and diagnosis models for low-prevalence diseases, where limited labeled data would otherwise hinder training.

Recent research has scaled ECG model and dataset sizes to enhance representation learning, with quite diverse choices of architectures and paradigms. As we show in \Cref{fig:size_performance}, the current landscape contains models trained using self-supervised (SSL) and supervised learning (SL), mainly based on transformers and convolutional neural networks (CNN), most notably the ResNet architecture~\citep{he2016deep}. 
For instance, ECG-FM~\citep{mckeen2024ecg} scales a transformer-based model to $90.9\,\textrm{M}$ parameters under SSL, while ECGFounder~\citep{li2024electrocardiogram}, a CNN-based model, scales the supervised training set to over $10\,\textrm{M}$ annotated ECG records.

In the literature, scaling laws~\citep{kaplan2020scaling, brandfonbrener2024loss, hernandez2021scaling} characterize the predictable gains in model performance relative to model size, dataset size, and total compute.
However, recent benchmarking \citep{al2025benchmarking} and large-scale hyperparameter optimization \citep{nolin2026foundation} of ECG foundation models suggest that \emph{clear scaling is not yet evident in this domain}.
In fact, \cite{al2025benchmarking} introduced a lightweight model with just $3.8\,\textrm{M}$ parameters that matches or outperforms pretrained models over $20$ times its size.
On the other hand, \cite{coppola2024hubert} showed that a $30\,\textrm{M}$-parameter variant of HuBERT achieves superior downstream fine-tuning performance compared to its significantly larger $188\,\textrm{M}$ counterpart.
\cite{al2025benchmarking} concluded that neither increased model size nor larger data scales consistently translate into superior transfer performance, revealing a distinct non-monotonicity in ECG scaling regimes. Similar conclusions were reached  for electroencephalograms (EEG)~\citep{yang2026eeg}.
Experiment design often complicates these comparisons. For example, \citet{nolin2026foundation} observed that SSL outperforms SL when diagnosing new conditions, yet their SL baseline utilized only $1.51\,\text{M}$ parameters, making it significantly smaller than its SSL counterpart. 
We argue that \textbf{\it this unpredictability arises because existing empirical observations do not clearly and systematically disentangle the effects of architectural inductive biases, hyperparameters, and pre-training paradigm}.

To address this problem, we systematically investigate the scaling behavior of ECG models by pre-training a suite of over 120 models, ranging from $20\,\textrm{K}$ to $200\,\textrm{M}$ parameters, on CODE~\citep{ribeiro2021code}, currently the second-largest public ECG dataset ($2.3\,\textrm{M}$ records, $1.6\,\textrm{M}$ patients).
We further conduct experiments on MIMIC-IV~\citep{PhysioNet-mimic-iv-ecg-1.0} ($0.8\,\textrm{M}$ records, $927$ labels) to study the effect of label scaling.
Our study considers four representative ECG configurations to disentangle the effects of architectural inductive bias and pre-training paradigms: ResNet-SL, ResNet-SSL (via contrastive learning), Transformer-SL, and Transformer-SSL (via masked autoencoding). 
By evaluating these models across both in-distribution (ID) and out-of-distribution (OOD) scenarios, we found that the scaling effect is strictly dependent on the coupling between architecture and paradigm:

\begin{figure}[t]
  \centering
  \includegraphics[width=1.0\textwidth]{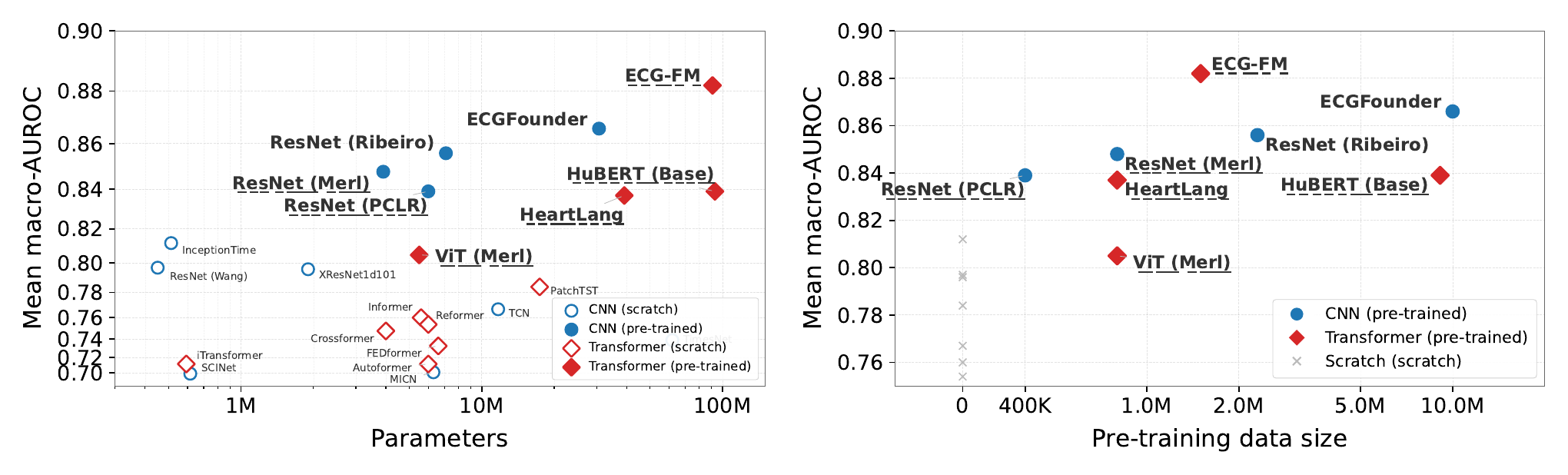}
\caption{Average downstream performance across 10 datasets (see Table \ref{tab:benchmark} for full results). \dotuline{Underlined methods} indicate self-supervised foundation models. The results do not show a clear scaling trend; instead, scaling behaviors diverge across different architectures and training paradigms. }
  \label{fig:size_performance}
  \vspace{-3mm}
\end{figure}

\begin{enumerate} 
\setlength{\leftskip}{-2.1em}
\item  In the ID scenario, SL models are primarily bottlenecked by dataset size rather than model size, whereas SSL models show no immediate evidence of scaling bottlenecks in either.

\item  OOD scaling exhibits an architectural bias, with ResNets yielding
$1.3$ to $2.5$ times higher parameter scaling efficiency than Transformers,
and a paradigm bias, where SSL achieves up to $16$ times higher data
scaling efficiency than SL.

\item Across the observed scales, ResNet-based models generally achieve the lowest absolute OOD loss; in parallel, SSL exhibits superior transfer efficiency ($7.6\times$) and lower OOD loss on unseen clinical tasks compared to SL, and Transformer-SSL ultimately prevails at large model sizes.

\item  While scaling the label space in supervised pre-training improves ID performance, OOD generalization remains sensitive to both architecture and the specific choice of pre-training labels.
\end{enumerate}

\section{Related Work}
\textbf{Scaling laws.} 
Neural scaling laws~\citep{kaplan2020scaling} describe how model performance improves as a power law in model size, dataset size, and training compute. \citet{hoffmann2022training} further formulated a parametric scaling law bounded by an irreducible error, establishing that optimal scaling efficiency is achieved when model size and training data size grow proportionally.
These principles have been extended to the transfer setting, where pre-training and downstream distributions differ~\citep{hernandez2021scaling}.
Recently, \citet{brandfonbrener2024loss} introduced loss-to-loss prediction, demonstrating that shifted power laws relate losses from pre-training to downstream tasks.
Beyond the natural language domain, scaling laws have been extended to other domains~\citep{zhai2022scaling, yao2024towards, chen2025owls}. For example,
\citet{zhai2022scaling} characterized the performance--compute frontier for Vision Transformers, observing power-law scaling with a double-saturation effect at extreme compute.
In the speech domain, scaling behavior has been explored for self-supervised and supervised models, though primarily with respect to model size~\citep{chen2025owls}.
A separate line of work examines how architectural inductive biases interact with scaling.
\citet{tay2023scaling} conducted a systematic study, evaluating ten architectures and showing that the best-performing model can change across scales, and that upstream gains do not always transfer downstream.
However, existing studies investigate architectures and pre-training paradigms in isolation.
To our knowledge, no prior work has jointly disentangled their coupled effects.

\textbf{ECG foundation models.}
The two dominant architectural families are convolutional networks, typically ResNets, whose local inductive biases align well with periodic signal morphology, and transformers, which can capture long-range temporal dependencies.
Under the supervised paradigm, scaling relies on large annotated datasets. For instance, ECGFounder \citep{li2024electrocardiogram} pre-trained a backbone on over $10$M recordings to predict $150$ labels, while AnyECG \citep{li2026anyecg} expanded this approach to $13.3$M ECGs across $1{,}172$ labels.
To reduce the reliance on annotated data, the self-supervised paradigm has become widely adopted, with both contrastive \citep{kiyasseh2021clocs, liu2024zero} and masked objectives \citep{jin2025reading, coppola2024hubert,gu2026cardiac} being extensively explored.
However, recent benchmarking studies raise concerns about the practical value of scaling data and ECG models. \citet{al2025benchmarking} and \citet{lunelli2025benchecg} evaluated ECG foundation models across clinical tasks and found that compact models with only a few million parameters can match or
outperform much larger models.
\citet{nolin2026foundation} directly compared SL and SSL ECG foundation models and observed that
SSL outperforms SL when diagnosing novel conditions with limited labels.
However, because these studies alter architectures and paradigms simultaneously, they cannot isolate the impact of any single factor.

\section{Preliminaries}

\textbf{Neural scaling laws.} We model the scaling behavior of both ID and OOD test loss as a function of  model size and data size, following the joint scaling law proposed by \cite{hoffmann2022training}
\footnote{The training loss studied by \citet{hoffmann2022training} serves as an unbiased estimate of the test loss.}:
\begin{equation}
    L(N, D) = E + A\cdot{N^{-\alpha}} + B\cdot{D^{-\beta}},
    \label{eq:joint_scaling}
\end{equation}
where $L$ is the test loss, $N$ is the number of parameters, $D$ is the number of pre-training samples, $E \geq 0$ is the irreducible loss floor, and $\alpha,\beta \geq 0$ are the scaling exponents for model size and dataset size, respectively. 
Letting the number of data samples or learnable parameters go to infinity, Eq.\eqref{eq:joint_scaling}~reduces to its marginal forms:
\begin{equation}
    L(N) = E + A\cdot{N^{-\alpha}}, \qquad L(D) = E + B\cdot{D^{-\beta}},
    \label{eq:marginal_scaling}
\end{equation}

where $\alpha$ and $\beta$ measure scaling efficiency.
When scaling either the model size or the dataset size by a factor of $k$, we define the excess loss reduction ratio for the marginal forms in Eq.~\eqref{eq:marginal_scaling} as:
\begin{equation}
    \frac{\Delta L(kN)}{\Delta L(N)}
    = k^{-\alpha}, \qquad
    \frac{\Delta L(kD)}{\Delta L(D)}
    = k^{-\beta},
    \label{eq:excess_loss_reduction}
\end{equation}
where $\Delta L(N) = L(N) - E$ and $\Delta L(D) = L(D) - E$ denote the excess test loss.
A $k$-fold increase in $N$ (resp. $D$) reduces $\Delta L$ to $k^{-\alpha}$ (resp. $k^{-\beta}$) of its original value when there is no data (resp. model) size bottleneck.
Hence, larger values of $\alpha$ or $\beta$ indicate higher
scaling efficiencies \footnote{For example, with a 10-fold increase in the number of learnable parameters ($k=10$), a coefficient $\alpha=0.3$ implies halving the excess loss ($k^{-\alpha} \approx 0.50$), while $\alpha=0.1$ implies a more modest reduction to $k^{-\alpha} \approx 0.79$ of its original value.}.

\textbf{Loss-to-loss scaling laws.}
Following \citet{brandfonbrener2024loss}, we investigate whether the ID test loss $L_{\rm ID}$ can serve as a predictor of the OOD test loss $L_{\rm OOD}$ across downstream datasets, characterizing the transfer relationship via a power law in the excess ID test loss:
\begin{equation}
    \Delta L_{\rm OOD} \approx K \cdot \big( \Delta L_{\rm ID} \big)^{\kappa},
    \label{eq:loss_to_loss}
\end{equation}
where $\Delta L_{\rm OOD} = L_{\rm OOD}(\widehat{f}^{N,D}) - E_{\rm OOD}$ and $\Delta L_{\rm ID} = L_{\rm ID}(\widehat{f}^{N,D}) - E_{\rm ID}$ are the excess losses of a model $\widehat{f}^{N,D}$ trained on the ID dataset, and $E_{\rm ID}, E_{\rm OOD}$ are the irreducible loss floors from the joint scaling fits.
The coefficients $K > 0$ and $\kappa > 0$ are fitted per OOD dataset.
The transfer exponent $\kappa$ governs transfer efficiency: $\kappa{=}1$ implies linear transfer, $\kappa{<}1$ indicates diminishing returns, and $\kappa{>}1$ indicates accelerating returns.

\section{Methods}
\subsection{Model architectures and paradigms}

Given an ECG signal $\mathbf{X}\in\mathbb{R}^{T\times C}$, where $T$ denotes the signal length and $C$ the number of leads (channels), and diagnostic binary labels $\mathbf{y}\in\mathbb{R}^{Y}$, where $Y$ is the number of labels, we describe the pre-training paradigms and the two model architectures.

For ResNet, we adopt the 1D ResNet architecture proposed by \citet{ribeiro2020automatic}, which adapts the vanilla ResNet design \citep{he2016deep} to ECG signals through minor architectural modifications. 
For the Transformer, we adopt HeartLang~\citep{jin2025reading}, which keeps a vanilla Transformer as the backbone model, supports both SL and SSL pre-training paradigms, and involves minor adaptations for ECG scenarios.

\textbf{ResNet-SL.} 
Without loss of generality, following the formulation in \citet{tan2019efficientnet}, we represent a $\widehat{f}_{\operatorname{ResNet}}$ as a composition of $s$ residual blocks $\{\mathcal{F}^{(i)}\}_{i=1}^{s}$, where each block includes a downsampling operation and a stack of convolutional layers with residual connections:
\begin{equation}
\widehat{f}_{\operatorname{ResNet}}(X) = \bigodot_{i=1}^s \mathcal{F}^{(i)} \big(\mathbf{Z}^{(i-1)}_{\langle H_i, C_i\rangle}\big),
\end{equation}
where $\mathbf{Z}^{(i-1)}$ denotes the input tensor to $\mathcal{F}^{(i)}$ with shape $\langle H_i, C_i \rangle$, with $\mathbf{Z}^{(0)}=\mathbf{X}$.
Here, $H_i$ and $C_i$ correspond to the sequence length and channel dimension, respectively.
The temporal dimension $H_i$ typically decreases across blocks due to downsampling operations, while the channel dimension $C_i$ increases to enhance representational capacity.
The output layer is a linear classifier that maps the output of the final block to the label space, i.e., $\widehat{\mathbf{y}} = \operatorname{Linear}(\mathbf{Z}^{(s)})$.

\textbf{ResNet-SSL.}
The SSL variant replaces the SL classifier with a two-layer MLP, projecting backbone representations to a 128-dimensional space for contrastive learning.
We adopt the Contrastive Multi-Segment Coding (CMSC) pre-training  objective from \citet{kiyasseh2021clocs}, which is also adopted by 
ECG-FM~\citep{mckeen2024ecg}. 
Each ECG is partitioned into two adjacent, non-overlapping segments $(\mathbf{X}_1, \mathbf{X}_2)$ to form a positive pair. 
Records from other patients within the same batch serve as negative samples. The model maximizes agreement between positive pairs while minimizing similarity to negatives. 
The projection head is later replaced by a linear classifier for evaluation.

\textbf{Transformer-SSL.}
The SSL strategy of HeartLang is formulated as a masked reconstruction task, analogous to masked language modeling.
Given the ECG signal $\mathbf{X}$, the HeartLang tokenizer segments it into a sequence of $l$ heartbeat segments $\{\mathbf{b}^{(i)}\}_{i=1}^{l}$, where each $\mathbf{b}^{(i)} \in \mathbb{R}^{K \times C}$ is padded or truncated to a fixed temporal length $K$.
To construct training targets, the unmasked heartbeat segment is discretized using an off-the-shelf ECG codebook $\mathcal{V}$, which assigns a discrete code $v_i \in \mathcal{V}$ to each heartbeat $\mathbf{b}^{(i)}$. A binary mask vector $\mathbf{m} \in \{0,1\}^{l}$ is then sampled, and masked positions are replaced with a learnable mask token $\mathbf{e}_M$, resulting in the masked input sequence
\begin{equation}
\mathcal{X}_M = \{ \mathbf{b}^{(i)} \;\text{if}\; \mathbf{m}[i] = 0,\; \text{else}\; \mathbf{e}_M \}.
\end{equation}
The masked sequence is encoded by a Transformer backbone to produce latent representations $\mathbf{H} = \widehat{f}_{\operatorname{Transformer}}(\mathcal{X}_M)$, where $\mathbf{H} \in \mathbb{R}^{l \times D}$ and $D$ is the hidden dimension of the transformer. 
A linear prediction head is applied to produce logits over the codebook as
$\mathbf{P}=\operatorname{Softmax}(\operatorname{Linear}(\mathbf{H}))$, where $\mathbf{P}\in\mathbb{R}^{l\times|\mathcal V|}$.
The training objective is the cross-entropy computed only on masked positions.

\textbf{Transformer-SL.}
In the SL adaptation, \cite{jin2025reading} retain the HeartLang tokenizer but shift from generative reconstruction to direct classification.
Similar to the Vision Transformer, the Transformer backbone processes the full sequence $\{\mathbf{b}^{(i)}\}_{i=1}^{l}$, prepending a special class token $[\text{CLS}]$ to aggregate global information. The linear output layer is applied to classify:
\begin{equation}
\widehat{\mathbf{y}} = \text{Softmax}(\text{Linear}(\mathbf{h}_{\text{cls}})), \quad \mathbf{H} = \widehat{f}_{\operatorname{Transformer}}([\mathrm{CLS}],\{\mathbf{b}^{(i)}\}_{i=1}^{l}).
\end{equation}

\subsection{Scaling strategies} 
\label{sec:scaling_strategies}
\textbf{Parameter scaling.}
For each backbone architecture, we instantiate a family of model variants with sizes ranging from $10^{4}$ to $10^{8}$ parameters.
For the ResNet backbone, we adopt a compound scaling strategy inspired by \citet{tan2019efficientnet}. Given a baseline network parameterized by $\{\widehat{s}, \widehat{H}_i, \widehat{C}_i\}$, we construct scaled variants by uniformly adjusting network depth and width throughout the backbone:
\begin{equation}
\widehat{f}_{\operatorname{ResNet}}(d, w) = \bigodot_{i=1}^{d \cdot \widehat{s}} \widehat{\mathcal{F}}_{i}
\Big(\mathbf{Z}_{\langle \widehat{H}_i, \; w \cdot \widehat{C}_i \rangle}\Big),
\end{equation}
where $d$ and $w$ denote the depth and width coefficients, respectively, which are jointly scaled. 
To scale the Transformer, we jointly increase the number of layers ($N_{\text{layer}}$), the dimensionality of the feed-forward sublayers ($D_f$), the number of attention heads per layer ($N_{\text{head}}$), and the hidden dimension of the attention outputs ($D$). Detailed configurations are available in Appendix~\ref{sec:app_model_details}.

\textbf{Data scaling.}
We constructed several subsets of the CODE dataset by varying the proportion of patients used during pre-training from $0.001\%$ to $90\%$.\footnote{This range corresponds to record counts from $10^{3}$ to $10^{6}$.}
To save compute, we sampled a few model sizes per data scale ensuring good joint fits of Eq.~\eqref{eq:joint_scaling} in-distribution, achieving a goodness-of-fit $R^2>0.95$. The $10^7$-parameter model was fully evaluated across all data scales, as it represents the largest scale where neither model bottlenecks nor ID overfitting were observed.

\textbf{Compute scaling.}
The total training compute budget is defined as $C = 6 \times \text{MACs}_{\text{fwd}} \times B \times S$,
where $\text{MACs}_{\text{fwd}}$ denotes the number of Multiply-ACcumulate operations (MACs) for a single forward pass on one input sample (measured using
\texttt{fvcore}~\citep{fvcore2026}),
$B$ is the batch size, and $S$ is the total number of gradient steps. The factor of $6$ decomposes as $2 \times 3$: the factor of $2$ converts MACs to FLOPs (each MAC comprises one multiplication and one
addition), and the factor of $3$ accounts for the backward pass being approximately twice as expensive as the forward pass~\citep{kaplan2020scaling}.

\textbf{Label scaling.}
We partitioned the MIMIC-IV dataset into training (160K records), validation (40K records), and test (20K records) sets. To investigate the impact of label scaling on model performance, we considered two scenarios: (i)~Dominant Classes: we fixed the 10\% most frequent classes ($|\mathcal{C}_{\text{most}}|=92$), with a maximum prevalence of 22.67\%, and randomly expanded the training label set to 30\%, 50\%, and 100\% of the total classes, while evaluating on $\mathcal{C}_{\text{most}}$ and CPSC2018. (ii)~Tail Classes: after excluding classes with fewer than 10 samples, we fixed the 2\% least frequent classes ($|\mathcal{C}_{\text{least}}|=9$, prevalence $\le 0.001\%$) and followed the same scaling strategy, testing on $\mathcal{C}_{\text{least}}$. 

\subsection{Pre-training and evaluation setup}
\label{sec:pretraining_setup}

\textbf{Pre-training datasets.}
We pretrain our models on CODE \citep{ribeiro2020automatic}, one of the largest 12-lead ECG datasets, comprising over $1.6\,\textrm{M}$ patients and $2.3\,\textrm{M}$ ECG recordings. The dataset provides labels for six relatively balanced and expert-annotated labels, covering both atrial and ventricular rhythm disorders as well as conduction abnormalities. 
To investigate label scaling, we pre-train separate models on a subset of MIMIC-IV ECG \citep{PhysioNet-mimic-iv-ecg-1.0} containing $160\,\textrm{K}$ records from $32\,\textrm{K}$ patients, annotated with $927$ long-tailed ICD-10 cardiac labels.

\textbf{OOD datasets.} We selected eleven datasets for OOD evaluation: CPSC2018~\citep{liu2018open}, Chapman-Shaoxing-Ningbo (CSN)~\citep{PhysioNet-ecg-arrhythmia-1.0.0}, EchoNext~\citep{PhysioNet-echonext-1.1.0}, PTB-XL (five subsets)~\citep{wagner2020ptb}, and MIMIC-IV (three subsets)~\citep{PhysioNet-mimiciv-3.1}. 
These datasets cover diverse clinical tasks, spanning adult ECG interpretation, structural heart disease detection, acute care prediction, and patient characteristic identification.
Dataset properties, preprocessing, and splitting strategies are provided in the Appendix~\ref{supplementary:dataset}.

\begin{figure}[t]
    \centering
    \includegraphics[width=1.0\linewidth]{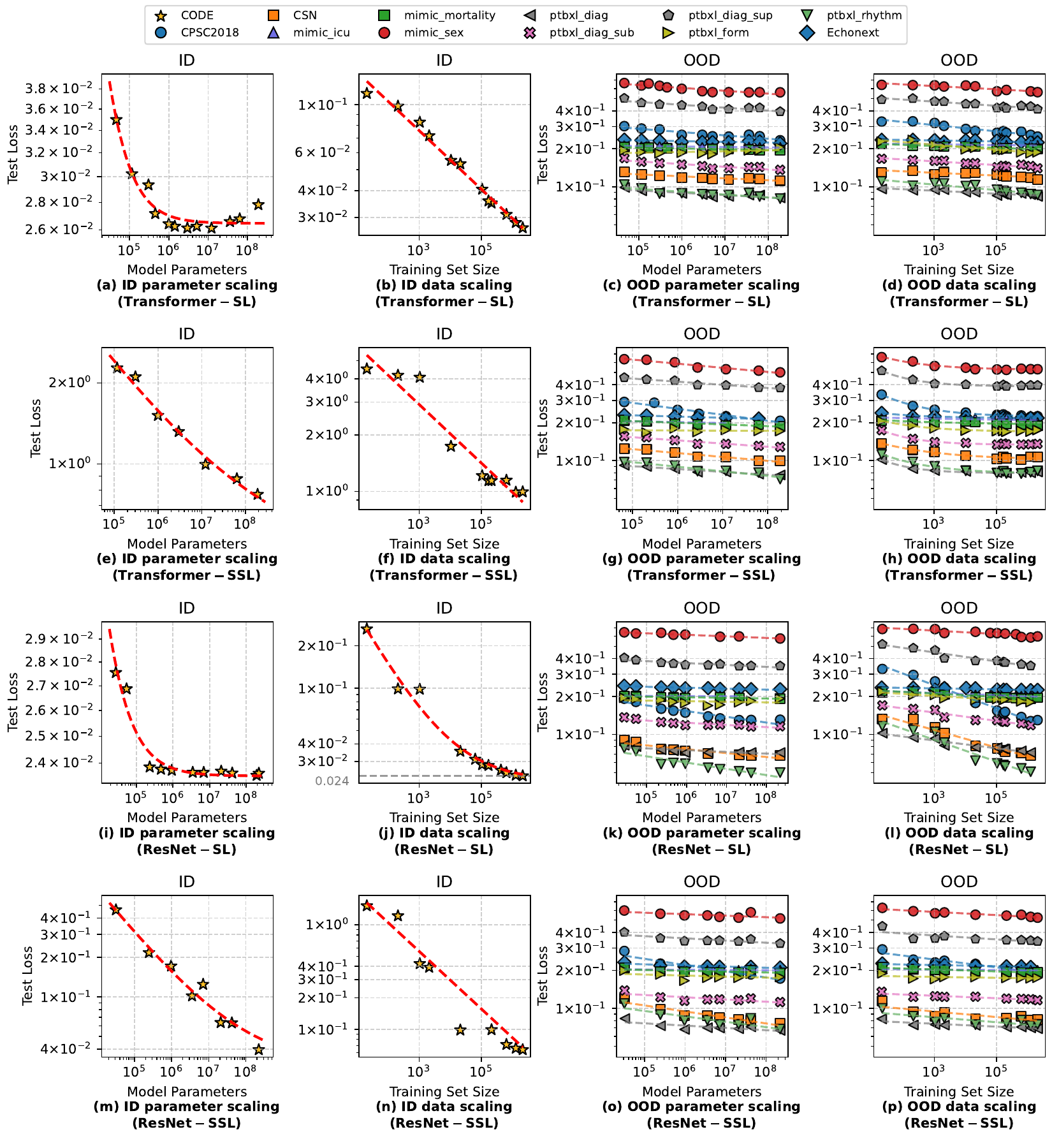}
    \vspace{-3mm}
    \caption{ID and OOD results of Transformer-SL, Transformer-SSL, ResNet-SL, and ResNet-SSL. Given a specific model or data scale, we report the optimal metric across all pre-training steps. The dashed line represents a least-squares fit based on Eq.~\eqref{eq:marginal_scaling}. Full table is available in Appendix~\ref{appendix:parameter_data_scaling}.}
    \label{fig:param_data_scaling}
\end{figure}

\textbf{Pre-training details.} We pretrain all models on CODE for a fixed budget of $1.5 \times 10^6$ optimization steps\footnote{We define one step as a model update corresponding to a batch size of 128}. 
The model performance is evaluated every $3{,}000$ steps on both ID and OOD scenarios, where ID corresponds to a consistent held-out 10\% test subset of CODE for fair comparison.
The maximum learning rate is $10^{-3}$ or $10^{-4}$, depending on the model size, with a linear warm-up over $10^4$ training steps, followed by cosine decay for the remaining steps.
For supervised pre-training, we use the cross-entropy loss and the Adam optimizer. Training is terminated earlier if sustained overfitting is observed.
For self-supervised pre-training, we employ cross-entropy loss for the masked learning objective and a contrastive loss~\citep{kiyasseh2021clocs} for the contrastive approach.
Setup details for the label scaling experiment on MIMIC-IV are provided in Appendix~\ref{sec:supp_pretraining_eval_setup}.

\textbf{Evaluation details.}
We report the best results across training steps.
ID performance is evaluated by computing the loss on a held-out test set using frozen model weights.
For OOD evaluation, we perform linear probing on the learned representations by training a single-layer linear classifier.
This classifier is trained on the training split of each OOD dataset, using early stopping based on validation performance. The final performance is reported on the test set. Details are provided in Appendix~\ref{sec:supp_pretraining_eval_setup}.

\section{Parameter Scaling and Data Scaling}

As shown in Figure~\ref{fig:param_data_scaling}, we fit Eq.~\eqref{eq:marginal_scaling} to each dataset to quantify the ID and OOD losses as a function of model size $N$ and pre-training dataset size $D$.
The fitted parameters are reported in Figure~\ref{fig:heatmap}.

\textbf{ID evaluation.}
When it comes to the ID evaluation, SL models exhibit a clear saturation in parameter scaling. The SL curve (Figures~\ref{fig:param_data_scaling} (a,i)) bends sharply ($\alpha_{\text{ID}}{=}0.943/0.912$ for Transformer/ResNet) and flattens beyond ${\sim}10^6$ parameters as the loss approaches the estimated floor $E$, indicating a data bottleneck. 
By contrast, the data scaling curves
(Figures~\ref{fig:param_data_scaling} (b,j)) remain far from saturation ($\beta_{\text{ID}}{=}0.129/0.39$), confirming that data size is the main bottleneck.  
SSL models exhibit the opposite trend: scaling curves show no visible saturation (Figs.~\ref{fig:param_data_scaling} (e,f,m,n)), mirroring the irreducible loss-free power laws of \citet{kaplan2020scaling}.
Unlike SL objectives that can be solved with moderate capacity, SSL provides a richer objective that rewards increased model and data sizes.

\begin{tcolorbox}
[boxsep=0mm,left=1.2mm,right=1.2mm,colframe=black!55,colback=black!5]
{\textbf{Takeaway 1.} For ID evaluation, SL models are primarily restricted by dataset size rather than model size, whereas SSL models generally exhibit unsaturated scaling across both.}
\end{tcolorbox}

\textbf{OOD evaluation.}
All scaling exponents ($\alpha, \beta$) are positive, indicating that OOD performance improves when independently increasing either model capacity or pre-training data scale.
In parameter scaling, ResNet models demonstrate a clear advantage over Transformers, with mean scaling exponents $\alpha$ that are $1.3 \times$ and $2.5 \times$ larger in SL and SSL setups, respectively. This implies that to match the excess loss reduction achieved by a $10$-fold increase in ResNet parameters, a Transformer would require significantly more scaling: a $19$-fold increase for SL and a $288$-fold increase for SSL (see Eq.~\eqref{eq:excess_loss_reduction}).
For data scaling the paradigm gap dominates. Under SSL, the mean of exponents $\beta$ is $16 \times$ (Transformers) and $4 \times$ (ResNets) larger than their SL counterparts, suggesting that SSL consistently improves OOD data scaling efficiency across both architectures.

\begin{tcolorbox}
[boxsep=0mm,left=1.2mm,right=1.2mm,colframe=black!55,colback=black!5]
{\textbf{Takeaway 2.} OOD scaling is shaped by two distinct dimensions of inductive bias: an architectural bias where ResNets yield a $1.3 \times$ to $2.5 \times$  higher parameter efficiency ($\alpha$) than Transformers, and a paradigm bias where SSL provides up to $16 \times$  gain in data efficiency ($\beta$) over SL.}
\end{tcolorbox}

\begin{figure}[t]
    \centering
    \includegraphics[width=1.0\linewidth]{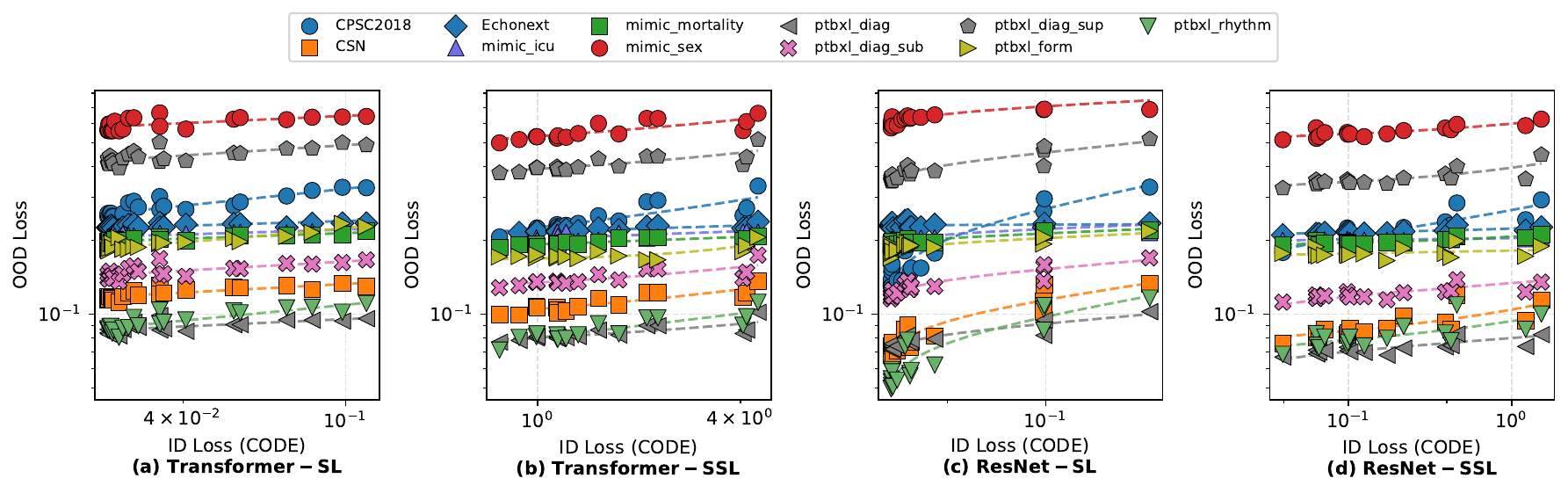}
    \caption{Loss-to-loss scaling curves for the four architecture-paradigm combinations. The dashed lines represent least-squares fits based on Eq.~\eqref{eq:loss_to_loss}. The full results are provided in Table~\ref{tab:loss2loss_joint}.}
    \label{fig:loss2loss}
\end{figure}

\textbf{ID-to-OOD scaling.}
As shown in Figure~\ref{fig:loss2loss}, we model the $L_{\rm ID}$-to-$L_{\rm OOD}$ scaling law by fitting Eq.~\eqref{eq:loss_to_loss} to paired $(L_{\rm ID}, L_{\rm OOD})$ observations. 
Recall that $\kappa$ denotes the transfer exponent governing how improvements in excess ID loss ($\Delta L_{\rm ID}$) translate into OOD gains ($\Delta L_{\rm OOD}$).
A transfer exponent of $\kappa < 1$ across nearly all conditions confirms that distribution shift consistently introduces a transfer bottleneck.
SSL models exhibit higher average transfer exponents (mean $\kappa_{\rm SSL} = 0.466$ and $0.270$ for Transformer and ResNet, respectively) than their SL counterparts ($0.128$ and $0.133$). 
Unlike SSL, SL models operate near their ID loss floor where the residual excess loss $\Delta L_{\rm ID}$ is minimal, causing further ID improvements to yield only diminishing returns for OOD generalization.
Empirically, this efficiency gap is especially pronounced for unseen diagnostic tasks\footnote{Pre-training involves adult ECG interpretation, as categorized by \citet{al2025benchmarking}.} such as structural heart disease detection (EchoNext).
The Transformer-SSL achieves $\kappa{=}0.339$, compared to only $0.050$ for the Transformer-SL.  Similarly, patient characterization tasks (e.g., sex prediction) demonstrate the largest advantage for ResNet-SSL over ResNet-SL, yielding a $7.6$ times higher $\kappa$.

\begin{tcolorbox}
[boxsep=0mm,left=1.2mm,right=1.2mm,colframe=black!55,colback=black!5]
{\textbf{Takeaway 3:} SSL consistently outperforms SL in transfer efficiency ($\kappa$), with the advantage reaching up to $7.6\times$ in unseen tasks like sex prediction and structural heart disease detection. 
}
\end{tcolorbox}

\textbf{Analysis of absolute performance.}
\begin{figure}[t]
    \centering
    \includegraphics[width=1.0\linewidth]{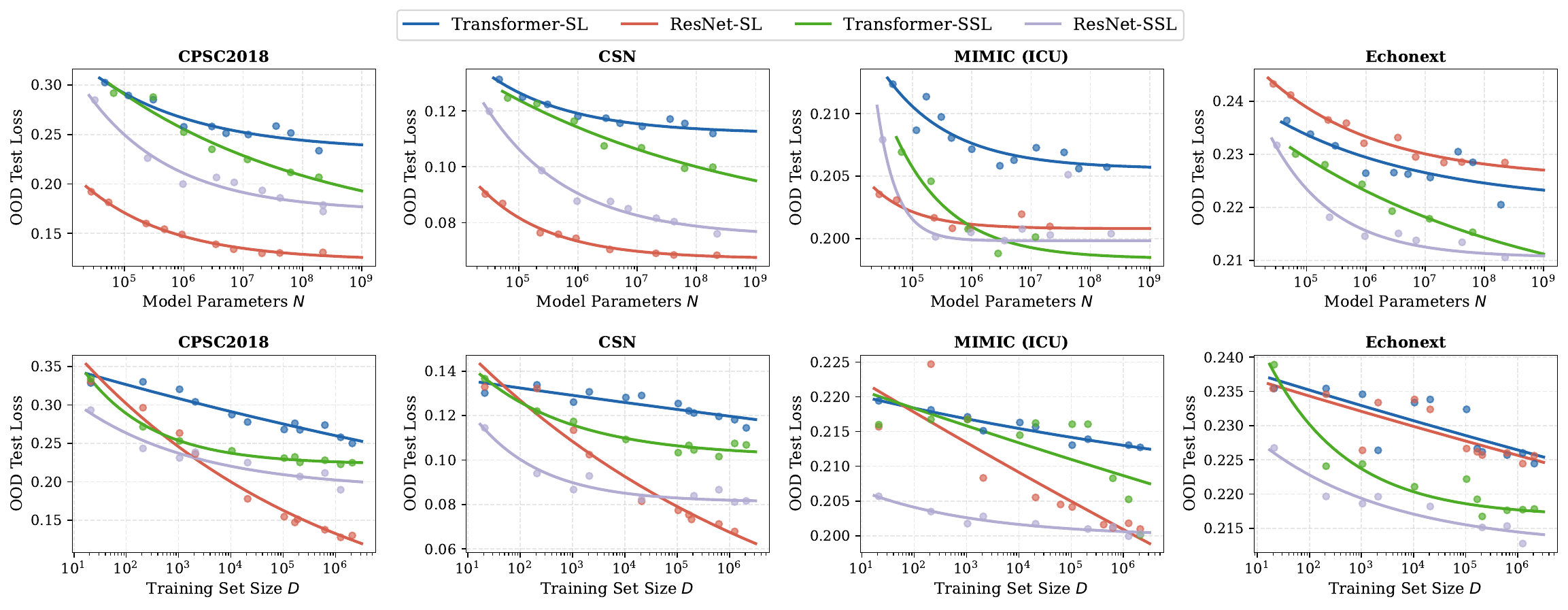}
    \caption{Scaling behavior for four architectural paradigms. Top and bottom rows illustrate OOD parameter and data scaling, respectively. Full results are provided in Appendix~\ref{sec:supplementary_crossover}}
    \label{fig:ood_scaling_crossover_2x4}
\end{figure}
Figure~\ref{fig:ood_scaling_crossover_2x4} reveals that ResNet-based models consistently dominate across most model and data scales regardless of the pre-training paradigm, highlighting a strong architectural inductive bias.
Besides, Transformer-SSL models consistently outperform their SL counterparts across all observed scales.
For unseen clinical tasks, such as structural heart disease detection, SSL consistently achieves lower test loss than SL within the observed scaling range.
Transformer-SSL only outperforms ResNet-SSL at large model sizes, which is consistent with our finding that Transformers are less parameter-efficient and with our benchmarking results in Section~\ref{sec:benchmark}.

\begin{tcolorbox}
[boxsep=0mm,left=1.2mm,right=1.2mm,colframe=black!55,colback=black!5]
{\textbf{Takeaway 4.} ResNet-based models generally achieve the lowest values of absolute OOD loss, suggesting an architectural inductive bias. For unseen clinical tasks, SSL consistently outperforms SL within the observed scales, with Transformer-SSL ultimately prevailing at large model sizes.}
\end{tcolorbox}

\begin{figure}[t]
    \centering
    \includegraphics[width=1.0\linewidth]{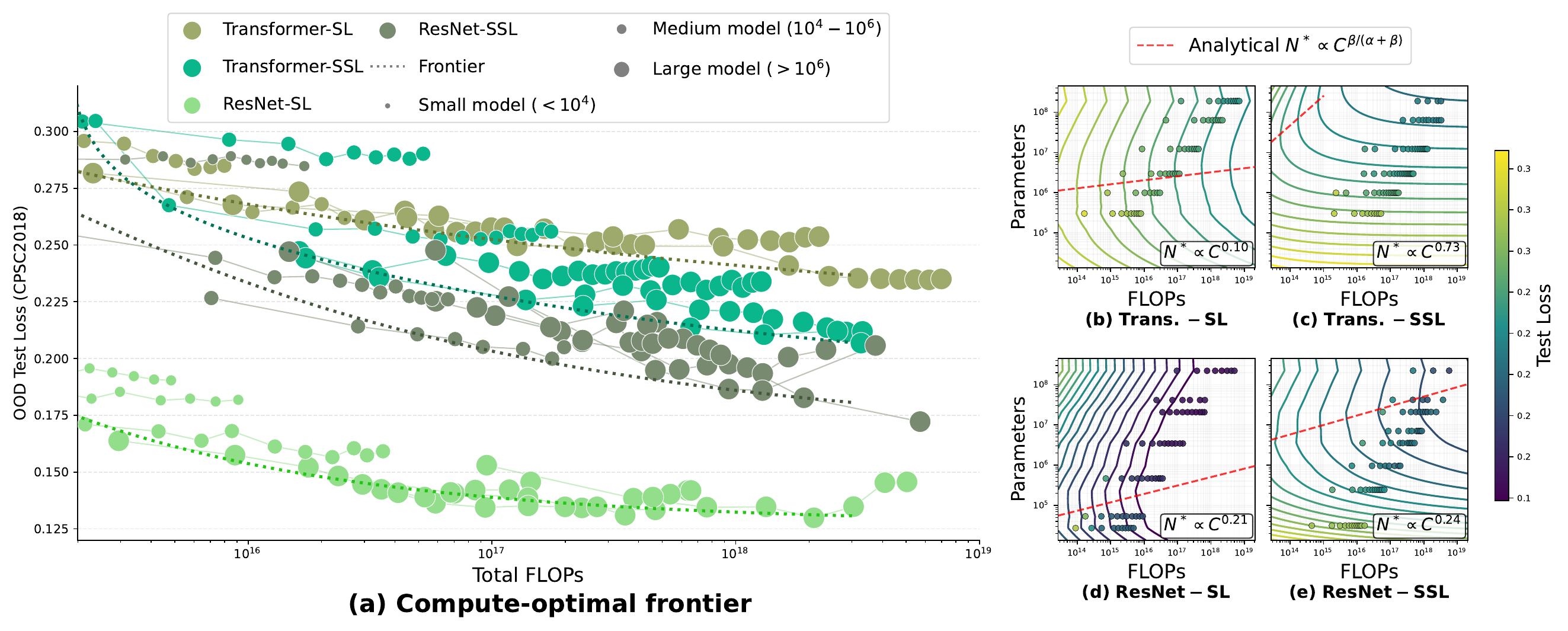}
    \caption{OOD evaluation on CPSC2018. (a)~Empirical frontier of test loss versus pre-training FLOPs.
    (b–e)~Iso-loss contours in the (C, N) plane, with the red dashed line indicating the compute-optimal model size $N^* \propto C^{\beta/(\alpha+\beta)}$. 
    Each panel reports the exponent of $N^*(C)$, indicating the fraction of marginal compute directed toward model size. 
    Full OOD results and derivation are in Appendix~\ref{sec:appendix_compute}.}
    \label{fig:compute_scaling_frontier}
\end{figure}

\textbf{Compute-optimal allocation.}
The Pareto frontier in Figure~\ref{fig:compute_scaling_frontier}(a) illustrates the compute-optimal allocation for each architecture-paradigm combination.
ResNet-SL yields the minimum test loss per FLOP, with ResNet-SSL following. 
Moreover, Transformer-SSL consistently outperforms Transformer-SL at equivalent FLOPs.
A natural follow-up question is how to allocate a fixed compute budget between model size and training data. 
Derived from the Chinchilla scaling law in Eq.~\eqref{eq:joint_scaling}, the compute-optimal allocations follow $N^* \propto C^{\beta/(\alpha+\beta)}$ and $D^*(C) \propto C^{\alpha/(\alpha+\beta)}$. As shown in Figure~\ref{fig:compute_scaling_frontier}(b-e), a distinct paradigm bias emerges: under the optimal allocation, Transformer-SL assigns 90\% of marginal compute to data (79\% for ResNet-SL), while Transformer-SSL inverts this pattern, directing 73\% toward model size. These trends hold across the remaining OOD datasets (Figure~\ref{fig:optimal_allocation_10x4}).

\begin{tcolorbox}
[boxsep=0mm,left=1.2mm,right=1.2mm,colframe=black!55,colback=black!5]
{\textbf{Takeaway 5:} Under compute-optimal allocation, ResNet-SL and Transformer-SL direct more than $70\%$ of marginal compute to data, while Transformer-SSL favors larger models over more data.}
\end{tcolorbox}

\label{subsec:label_scaling}
\begin{figure}[t]
    \centering
    \includegraphics[width=1.0\linewidth]{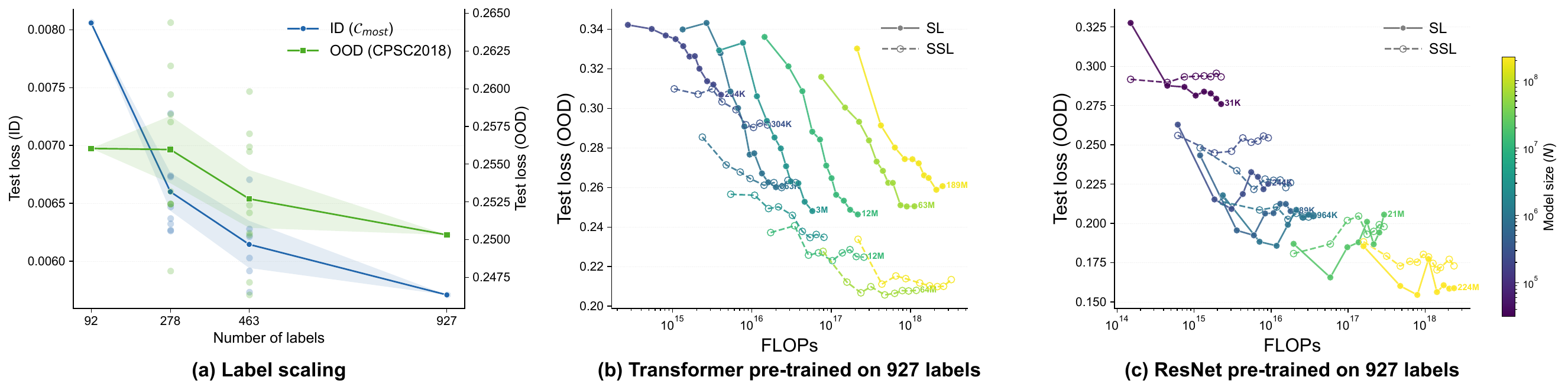}
    \caption{(a) Label scaling benefits ID performance, but OOD transfer depends on label selection and shows high variance. Results on $\mathcal{C}_{\text{least}}$ are shown in Figure~\ref{fig:id_label_scaling_1x4}. (b–c) On the CPSC2018 dataset, models of different sizes were evaluated during pre-training on 927 classes. Transformer-SSL outperforms Transformer-SL, while ResNet-SL and ResNet-SSL show comparable performance.}
    \label{fig:label_scaling}
\end{figure}

\textbf{The effect of scaling supervised labels.}
To study the effect of scaling supervised labels during pre-training, we pre-trained models on MIMIC-IV subsets ranging from 9 to 927 classes. As shown in Figure~\ref{fig:label_scaling}, increasing the number of pre-training labels consistently improves ID performance across both frequent and tail labels, indicating positive transfer.
While label scaling can improve average OOD performance, gains are highly variable and depend strongly on label composition. Under the 927-class regime, Transformer-SL fails to translate increased model capacity into stable OOD gains, performing worse than both ResNet-SL and Transformer-SSL. In contrast, ResNet-SL performs comparably to its SSL counterpart despite being trained with large labels. These results suggest that simply scaling supervised labels is suboptimal, highlighting the importance of architecture and pre-training paradigm choices.
Additional results are provided in Section~\ref{appendix:label_scaling}.

\section{Fine-tuning ECG Foundation Models}
\label{sec:benchmark}
As illustrated in Figure~\ref{fig:size_performance}, we evaluate state-of-the-art ECG foundation models against train-from-scratch baselines across 10 downstream tasks. Rather than exhibiting a uniform scaling law, the results reveal distinct performance trajectories influenced by architectural choice and pre-training paradigm.
Notably, CNN architectures, particularly ResNet variants, consistently outperform Transformer-based models at comparable model sizes, demonstrating architectural inductive bias and parameter efficiency.
The best overall performance is achieved by ECG-FM, which is the largest Transformer-based SSL model evaluated, followed by ECGFounder, a CNN-based SL model trained on the largest dataset.
This observation is consistent with our finding that Transformer-SSL is less parameter-efficient, but exhibits strong generalization on unseen clinical tasks at large model sizes.
Self-supervised models are more data-efficient than supervised models, and therefore models like ECG-FM and ResNet (Merl) can perform better than or comparably to ECGFounder and ResNet (Ribeiro), which were pre-trained on much larger ECG datasets.
Experiment details are available in Section~\ref{appendix:benchmark}.

\section{Conclusion and Future Work}
\label{sec:conclusion}

We systematically disentangle the effects of architecture and pre-training paradigms on scaling ECG models. Our findings yield several actionable insights: in ID scenarios, SL models are primarily limited by dataset size. Under OOD scenarios, ResNets show greater parameter efficiency than Transformers, and SSL is more data-efficient than SL. For unseen clinical tasks, SSL proves superior, offering higher transfer efficiency and lower OOD loss.
Scaling label size consistently improves the ID performance of SL models, but OOD generalization depends on the architecture and label choice. These results resolve conflicting conclusions in previous literature regarding why downstream performance does not always increase with larger models or more data.
Future work includes analyzing additional architectures, such as state-space models, and alternative pre-training paradigms, such as hybrid pre-training. Covering more diverse clinical tasks using private data could further enrich the analysis, 
though this remains beyond the scope of this study.

\section{Acknowledgments and Disclosure of Funding}
\label{sec:aknowledgment}
The authors are grateful to Elis Stefansson, Ziqi Zhang and Ziwei Luo for feedback on early versions of this manuscript.
Jiawei Li and Antônio Horta Ribeiro are financially supported by the eSSENCE and SciLifeLab, with the
project "Digital Biomarkers from the Electrocardiogram using Artificial Intelligence";
and, by the Wallenberg AI, Autonomous Systems and Software Program (WASP) funded by Knut and Alice Wallenberg Foundation. 
This research was partially supported by \emph{Kjell och M{\"a}rta Beijer Foundation}.
This project has received funding from the European Research Council (ERC) under the European Union's Horizon Europe research and innovation programme through grant agreement no. 101054643.
Computations were enabled by resources provided by the National Academic Infrastructure for Supercomputing in Sweden (NAISS), partially funded by the Swedish Research Council through grant agreement no. 2022-06725.

% ==== APPENDIX ====

\newpage
\appendix

\setcounter{equation}{0}
\renewcommand{\theequation}{S.\arabic{equation}}%
\setcounter{figure}{0}
\renewcommand{\thefigure}{S.\arabic{figure}}%
\setcounter{table}{0}
\renewcommand{\thetable}{S.\arabic{table}}%
\setcounter{section}{0}

\onecolumn
\pagenumbering{roman} 

\part{}

~ 
\vspace{10pt}

\parttoc

\setcounter{page}{0 }
\newpage

\section{Implementation Details}
\label{appendix:implementation_detail}
\subsection{Dataset characteristics and preprocessing details}
\label{supplementary:dataset}
\paragraph{Pre-training datasets.} The CODE dataset \citep{ribeiro2021code} provides annotations for six representative rhythm and conduction abnormalities, including first-degree atrioventricular block (1dAVb, $1.5\%$), right bundle branch block (RBBB, $2.7\%$), left bundle branch block (LBBB, $1.7\%$), sinus bradycardia (SB, $1.6\%$), atrial fibrillation (AF, $1.8\%$), and sinus tachycardia (ST, $2.1\%$). 
We do not pretrain on the Harvard–Emory ECG dataset, despite its larger cohort size, as it contains a large number of labels characterized by severe class imbalance, which could introduce unwanted long-tailed learning effects and complicate the analysis of architectural and paradigm inductive biases.
We evaluated label scaling using the MIMIC-IV dataset, which encompasses a vast array of clinical labels. 
To specifically investigate cardiac and cardiovascular conditions, we extracted $927$ ICD-10 codes starting with the letter 'I', which clinically denote diseases of the circulatory system.
Within this subset, the most frequently occurring diagnoses are I10 (essential primary hypertension, 22.67\%), I2510 (atherosclerotic heart disease, 15.90\%), I4891 (unspecified atrial fibrillation, 12.13\%), I509 (unspecified heart failure, 9.35\%), and I129 (hypertensive chronic kidney disease, 6.54\%).

\paragraph{OOD datasets.}
Table~\ref{tab:ood_datasets_summary} summarizes the general information and label sets of the OOD datasets.
Each dataset is partitioned into training, validation, and test sets, with splits strictly following the protocols established in prior studies; see Table~\ref{tab:data_split_preprocess} for details.
Notably, these datasets encompass distinct clinical tasks, such as adult ECG interpretation, demographic profiling, and structural heart disease detection. This diversity of tasks underscores the versatile diagnostic value of ECG signals. We adopt the categorization framework defined by \cite{al2025benchmarking} to organize these tasks, as summarized in Table~\ref{tab:task_categorization}.

\begin{table*}[b!]
\centering
\small
\setlength{\tabcolsep}{6pt}
\renewcommand{\arraystretch}{1.25}
\caption{External OOD evaluation datasets and their label spaces. ``/'' indicates that a single scalar patient count is not explicitly reported in the corresponding dataset description.}
\begin{tabular}{p{0.18\linewidth}rrp{0.56\linewidth}}
\hline
\textbf{Dataset} & \textbf{\#Patients} & \textbf{\#ECGs} & \textbf{Label space (representative)} \\
\hline
CPSC2018 \newline \citep{liu2018open} &
6{,}877 &
6{,}877 &
Nine single-label rhythm classes including normal sinus rhythm, atrial fibrillation, first-degree atrioventricular block, left bundle branch block, right bundle branch block, premature atrial contraction, premature ventricular contraction, ST-segment depression, and ST-segment elevation. \\
\hline
Chapman--Shaoxing--Ningbo (CSN)\newline~\citep{PhysioNet-ecg-arrhythmia-1.0.0} &
45{,}152 &
45{,}152 &
Electrocardiographic rhythm types and additional cardiovascular conditions encoded using SNOMED-CT terminology. The dataset contains eleven rhythm categories and sixty-seven additional diagnostic conditions; rhythm annotations are commonly grouped into sinus rhythm, sinus bradycardia, atrial fibrillation, and supraventricular tachycardia. \\
\hline
EchoNext\newline\citep{PhysioNet-echonext-1.1.0} &
/ &
100{,}000 &
Binary label for structural heart disease derived from paired echocardiography. A positive label indicates the presence of at least one of the following conditions: reduced left ventricular ejection fraction, increased left ventricular wall thickness, significant valvular heart disease, right ventricular dysfunction, pulmonary hypertension, or pericardial effusion. \\
\hline
PTB-XL\newline(Superclasses)\newline\citep{wagner2020ptb} &
18{,}869 &
21{,}799 &
Five high-level diagnostic categories including normal electrocardiogram, conduction disturbances, myocardial infarction, ventricular hypertrophy, and ST--T segment abnormalities. \\
\hline
PTB-XL \newline(Subclasses)\newline\citep{wagner2020ptb} &
18{,}869 &
21{,}799 &
Twenty-three fine-grained diagnostic categories covering atrioventricular and intraventricular conduction blocks, bundle branch blocks, pre-excitation syndromes, ventricular and atrial hypertrophy, and multiple myocardial infarction localizations. \\
\hline
PTB-XL\newline(Form)\newline\citep{wagner2020ptb} &
18{,}869 &
21{,}799 &
Nineteen electrocardiographic morphological statements describing waveform and interval abnormalities, including abnormal QRS complex, prolonged QT interval, pathological Q waves, low or high voltage patterns, ST-segment deviation, and T-wave inversion. \\
\hline
PTB-XL\newline(Rhythm)\newline\citep{wagner2020ptb} &
18{,}869 &
21{,}799 &
Twelve rhythm annotations including sinus rhythm, atrial fibrillation, atrial flutter, supraventricular tachycardia, sinus tachycardia, sinus bradycardia, paced rhythm, and other supraventricular or ventricular arrhythmias. \\
\hline
MIMIC (Sex)\newline\citep{PhysioNet-mimiciv-3.1} &
198{,}509 &
198{,}509 &
A binary label indicating patient sex (e.g., male or female). \\
\hline
MIMIC (Mortality)\newline\citep{PhysioNet-mimiciv-3.1} &
79{,}374 &
79{,}374 &
Six mortality labels are included, indicating whether death occurred within 1, 7, 28, 90, 180, or 365 days. \\
\hline
MIMIC (ICU)\newline\citep{PhysioNet-mimiciv-3.1} &
40{,}180 &
40{,}180 &
Two binary labels are included to indicate intensive care unit (ICU) admission: overall ICU stay and ICU admission within 24 hours. \\
\hline
\end{tabular}
\label{tab:ood_datasets_summary}
\end{table*}

\begin{table*}[h]
\centering
\small
\setlength{\tabcolsep}{6pt}
\renewcommand{\arraystretch}{1.2}
\caption{Dataset-specific data splits, preprocessing procedures, and code availability. All datasets are resampled to 100~Hz to align with the HeartLang codebook.}
\begin{tabular}{p{0.18\linewidth} p{0.22\linewidth} p{0.30\linewidth} p{0.20\linewidth}}
\hline
\textbf{Dataset name} & \textbf{Train / Validation / \newline Test split} & \textbf{Preprocessing techniques} & \textbf{Implementation \newline reference} \\
\hline
CODE &
90\% for pre-training, 10\% for ID evaluation &
baseline wander removal (high-pass filter); power-line interference attenuation (60~Hz) &
\cite{ribeiro2020automatic}\\
\hline
MIMIC-IV ECG \newline (pre-training data) &
25\% for pre-training, \newline 5\% for ID evaluation &
baseline wander removal (high-pass filter); power-line interference attenuation (60~Hz) &
\cite{al2025benchmarking} \\
\hline
CPSC2018 &
80\% for training, \newline 10\% for validation, \newline 10\% for testing &
standardization &
\cite{al2025benchmarking} \\
\hline
Chapman--Shaoxing--Ningbo (CSN) &
80\% for training, \newline 10\% for validation, \newline 10\% for testing  &
normalization, band-pass filtering &
\cite{jin2025reading} \\
\hline
EchoNext &
 80\% for training, \newline 10\% for validation, \newline 10\% for testing 
 &
 /
 &
\cite{al2025benchmarking} \\
\hline
PTB-XL &
80\% for training, \newline 10\% for validation, \newline 10\% for testing &
standardization
&
\cite{jin2025reading} \\
\hline
MIMIC-IV ECG \newline (OOD data) &
25\% for training, \newline 5\% for validation, \newline 10\% for testing 
& 
imputation, clipping
&
\cite{al2025benchmarking} \\
\hline
\end{tabular}
\label{tab:data_split_preprocess}
\end{table*}

\begin{table*}[h]
\centering
\small
\setlength{\tabcolsep}{6pt}
\renewcommand{\arraystretch}{1.2}
\caption{Categorization of evaluation datasets by clinical task domain~\citep{al2025benchmarking}.}
\begin{tabular}{p{0.22\linewidth} p{0.20\linewidth} p{0.45\linewidth}}
\hline
\textbf{Task domain} & \textbf{Dataset} & \textbf{Task description} \\
\hline
\multirow{4}{*}{\parbox{0.22\linewidth}{Adult ECG \newline interpretation}} 
& CODE & Multi-label rhythm and morphology classification \\
& CSN & Multi-label cardiac arrhythmia classification \\
& CPSC2018 & Multi-label arrhythmia classification \\
& PTB-XL & Multi-label diagnostic, rhythm, form, and subdiagnostic classification \\
\hline
Cardiac structure \newline \& function \newline (Structural heart disease\newline detection)
& EchoNext & Prediction of echocardiographic abnormalities from ECG \\
\hline
Acute care predictions 
& MIMIC (icu) & ICU admission prediction from emergency department ECG \\
& MIMIC (mortality) & In-hospital mortality prediction \\
\hline
Patient characteristics 
& MIMIC (sex) & Biological sex classification from ECG \\
\hline
\end{tabular}
\label{tab:task_categorization}
\end{table*}

\paragraph{Preprocessing details.}
As the off-the-shelf HeartLang codebook was originally developed on ECG signals sampled at 100~Hz, pre-training datasets are resampled to 100~Hz for consistency. Standard preprocessing is applied to suppress signal artifacts, including baseline wander removal using a high-pass filter and attenuation of power-line interference at 60~Hz. For OOD evaluation, datasets are likewise resampled to 100~Hz. We adopt distinct preprocessing strategies for each OOD dataset following established prior practices, as detailed in Table~\ref{tab:data_split_preprocess}.

\subsection{Model details}
\label{sec:app_model_details}
\begin{figure}[t]
    \centering
    \includegraphics[width=1.0\linewidth]{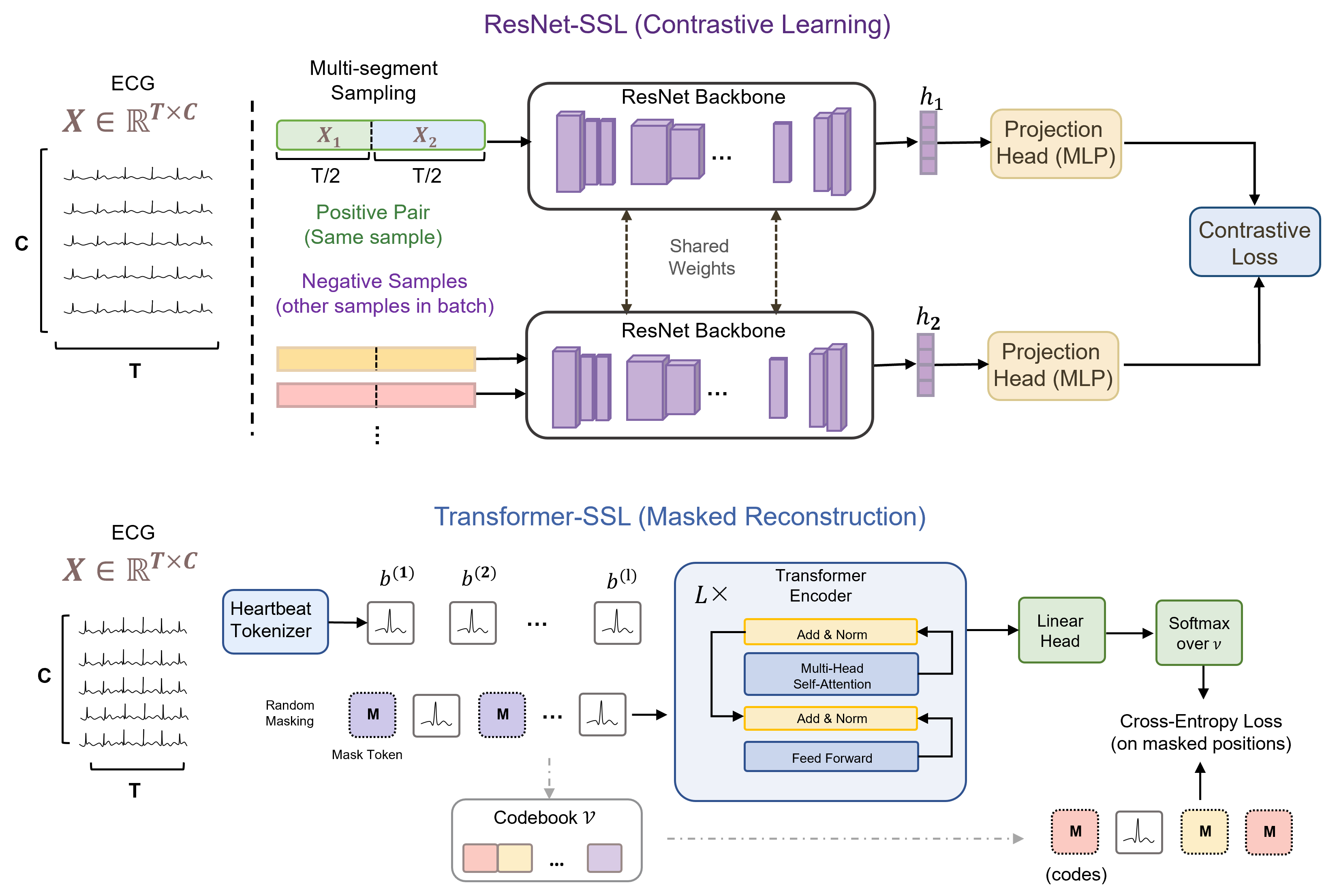}
    \caption{Self-supervised pre-training strategies for ResNet and Transformer. Both follow canonical paradigms to ensure broad representativeness without intricate pretext task overhead.}
    \label{fig:ssl_methodology}
\end{figure}

We provide full model configurations in Table~\ref{tab:model_variants}.
Additionally, we illustrate the SSL strategies used for ResNet and Transformer in Figure~\ref{fig:ssl_methodology}.

\paragraph{ResNet-SSL.}
We provide details of the contrastive learning objective proposed by \citet{kiyasseh2021clocs} and adopted in this paper.
Let $(\mathbf{X}_1^{(i)}, \mathbf{X}_2^{(i)})$ denote the two temporally adjacent, non-overlapping segments extracted from the $i$-th ECG in a mini-batch of size $B$.
Each segment is encoded by the backbone and projected to a 128-dimensional embedding via the MLP projection head $g(\cdot)$, yielding $\ell_2$-normalised representations $\mathbf{z}_1^{(i)}, \mathbf{z}_2^{(i)}$.
We concatenate all projections into a single set of $2B$ vectors $\{\mathbf{z}_k\}_{k=1}^{2B}$, where $\mathbf{z}_k = \mathbf{z}_1^{(k)}$ for $k \leq B$ and $\mathbf{z}_k = \mathbf{z}_2^{(k-B)}$ for $k > B$.
For each anchor $\mathbf{z}_k$, its positive set $\mathcal{P}(k)$ consists of all other embeddings in the batch derived from the same patient:
\begin{equation}
    \mathcal{P}(k) = \bigl\{j \neq k \mid \mathrm{pid}(j) = \mathrm{pid}(k)\bigr\},
\end{equation}
where $\mathrm{pid}(\cdot)$ maps an index to its patient identifier. This includes both the temporally adjacent segment from the same ECG and, when available, segments from other recordings of the same patient within the batch. All embeddings from different patients serve as negatives.
The training objective is the patient-specific loss:
\begin{equation}
    \mathcal{L}_{\mathrm{CMSC}} = -\frac{1}{2B} \sum_{k=1}^{2B}
    \frac{1}{|\mathcal{P}(k)|}\sum_{p \in \mathcal{P}(k)}
    \log \frac{
        \exp\!\left(\mathbf{z}_k^\top \mathbf{z}_p \,/\, \tau\right)
    }{
        \sum_{j \neq k} \exp\!\left(\mathbf{z}_k^\top \mathbf{z}_j \,/\, \tau\right)
    },
    \label{eq:cmsc_loss}
\end{equation}
where the denominator sums over all $2B - 1$ non-self entries in the batch, and $\tau$ is the temperature ($\tau = 0.1$).
After pre-training, the projection head is discarded and only the backbone 
is retained for downstream fine-tuning or linear probing.

\paragraph{Transformer.} 
In this study, we focus on encoder-only Transformer architectures. Prior empirical work~\citep{yao2024towards} suggests that, under comparable training setups, encoder-only and decoder-only Transformers exhibit similar parameter and data scaling behaviors.
For the SSL, we use the off-the-shelf HeartLang codebook~\citep{jin2025reading}, which was pre-trained on MIMIC-IV ($0.8M$ patients), and the random masking rate is $50\%$. 
Besides, it should be noted that the dataset size $D$ (number of training samples) is proportional to the total number of training tokens, since the HeartLang Tokenizer produces a fixed-length token sequence for each ECG input. As a consequence, fitting Eq.~\eqref{eq:joint_scaling} with $D$ measured in samples rather than tokens leaves $E$, $A$, $\alpha$, and $\beta$ unchanged.

\subsection{Pre-training and evaluation}
\label{sec:supp_pretraining_eval_setup}
\paragraph{Pre-training setup.}
For CODE dataset, the pretraining was conducted on a single NVIDIA A100 GPU (40GB) with batch size $B=128$. The maximum learning rate is set to $10^{-3}$ for models with fewer than $10^8$ parameters, and $10^{-4}$ for larger architectures. The pretraining is allowed to run for a maximum of $1.5 \times 10^6$ steps, with early stopping triggered if the ID validation loss fails to improve for at least $5$ consecutive epochs.
For the MIMIC-IV dataset \citep{PhysioNet-mimic-iv-ecg-1.0}, we pre-trained the models for a fixed 50 epochs ($1.2 \times 10^4$ steps) with a batch size of 128. The peak learning rate was set to $10^{-3}$ or $10^{-4}$ depending on the model size, with a linear warmup over the first epoch and cosine decay thereafter. To mitigate the severe class imbalance inherent in our long-tailed, multi-label setting, we employed focal loss as the objective function for SL. By default, we utilized Transformer-SL models with approximately $10^7$ parameters for ID evaluations.
Regarding optimization, all Transformer-SSL models were trained using AdamW, while ResNet-SSL models utilized the Adam optimizer, consistent with their original implementations.

\paragraph{Evaluation setup.} 
For the evaluation, we apply a linear classifier atop the frozen latent representations. This classifier is optimized via the Adam algorithm, configured with a learning rate of $10^{-3}$ and cross-entropy loss.  
The linear classifier is trained on the training split of each OOD dataset and validated on the corresponding validation split for a maximum of 200 epochs, with early stopping using a patience of 5 epochs. The final performance is reported on the held-out test set.
We opt for linear probing over full/partial fine-tuning to avoid the extensive hyperparameter search space, particularly for large-scale models. Empirical evidence~\citep{li2024electrocardiogram, coppola2024hubert, mckeen2024ecg} demonstrates that the fine-tuning performance of large models is sensitive to the choice of learning rates and freezing strategies.
The evaluation was conducted on either a single NVIDIA A40 GPU (48GB) or an A100 GPU (40GB), depending on the model size.

\begin{table*}[t]
\centering
\caption{Unified architectural configurations for ResNet and Transformer variants across parameter scales ($N$). Values in parentheses denote the specific hyperparameters for each architecture group.}
\label{tab:model_variants}
\small
\renewcommand{\arraystretch}{1.4}
\begin{tabular}{c p{0.32\linewidth} p{0.48\linewidth}}
\toprule
\textbf{Scale} ($N$) & \textbf{ResNet Config} $(d \cdot \widehat{s}, \; w \cdot \widehat{C}_i)$ & \textbf{Transformer Config} $(N_{\text{layer}}, \; D, \; N_{\text{head}}, \; D_f)$ \\
\midrule
$10^4$ & (3, 24); (3, 48) & (1, 24, 1, 96); (1, 48, 2, 192) \\
$10^5$ & (4, 64); (4, 90); (4, 128) & (1, 96, 2, 384); (2, 128, 4, 512) \\
$10^6$ & (5, 226); (5, 320) & (3, 192, 4, 768); (3, 256, 4, 1024); (4, 348, 4, 1392) \\
$10^7$ & (6, 452); (6, 640) & (5, 420, 8, 1680); (6, 512, 8, 2048) \\
$10^8$ & (7, 1024) & (8, 768, 8, 3072) \\
\bottomrule
\end{tabular}
\end{table*}

\section{Supplementary Results}
\label{sec:supplementary_result}

\subsection{Parameter scaling and data scaling} 
\label{appendix:parameter_data_scaling}
\begin{figure}[t!]
    \centering
    \includegraphics[width=0.95\linewidth]{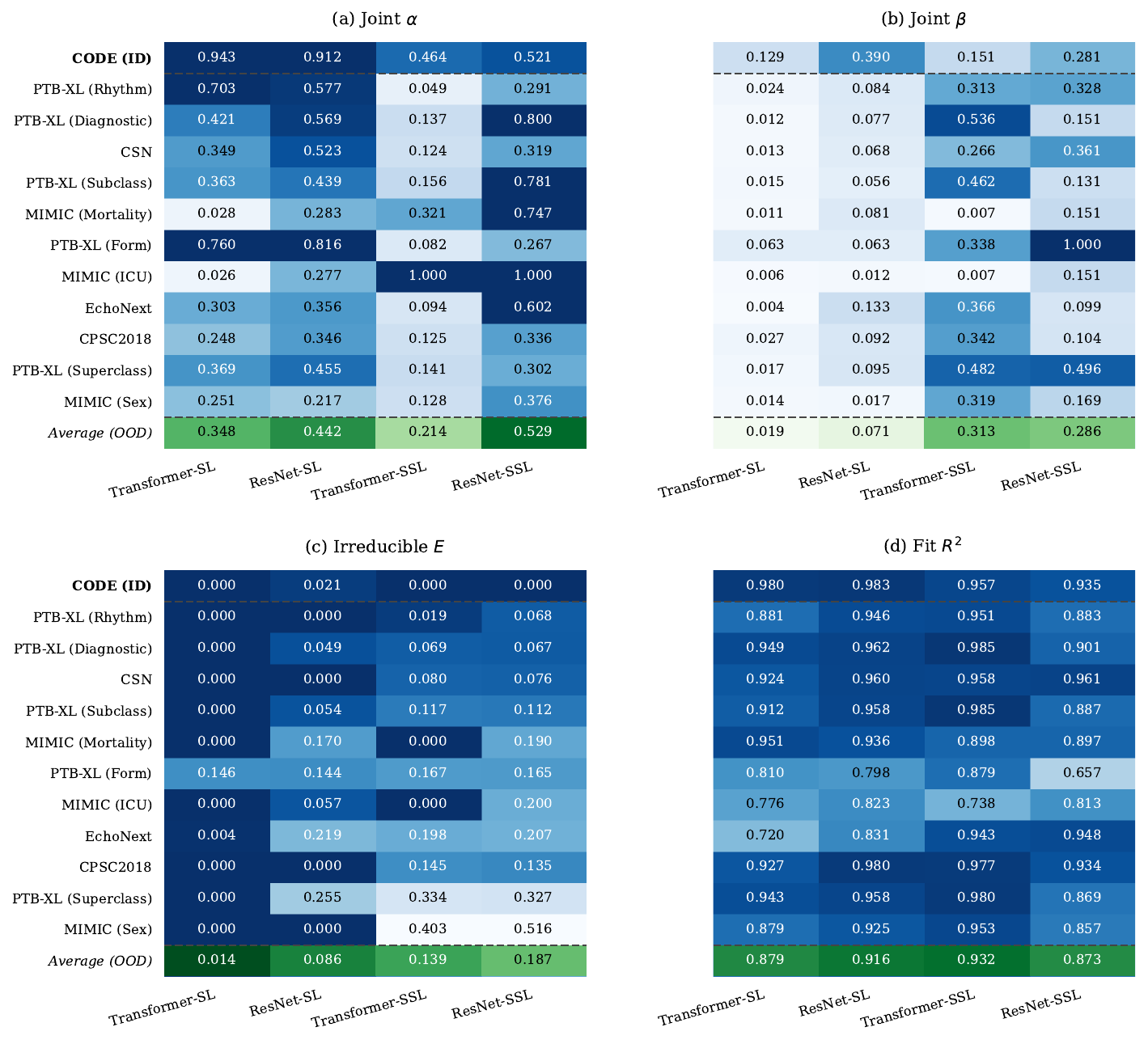}
    \caption{Fitted scaling exponents and goodness of fit ($R^2$) for the joint scaling laws (Eq.~\eqref{eq:joint_scaling}) across four architecture–paradigm combinations and all datasets.}
    \label{fig:heatmap}
\end{figure}

Figure~\ref{fig:heatmap} summarizes the fitted parameters of Eq.~\eqref{eq:joint_scaling} obtained by fitting to each architecture--paradigm combination and dataset. Panels~(a--c) report the joint scaling exponents $\alpha$, $\beta$, and the irreducible loss $E$, while panel~(d) reports the coefficient of determination $R^2$ of each fit. The majority of fits achieve $R^2{>}0.87$.
A salient pattern across OOD benchmarks is that SL models consistently exhibit $\alpha \gg \beta$. 
For instance, Transformer-SL models result in $\alpha{=}0.703$ and $\beta{=}0.024$ on PTB-XL (rhythm), and in $\alpha{=}0.349$ vs.\, $\beta{=}0.013$ on CSN, suggesting that supervised models exhibit higher parameter efficiency compared to data efficiency. 
Concerning SSL approaches,  Transformer-SSL results in $\beta{>}\alpha$ on most OOD datasets, suggesting that masked pre-training is data-hungry and benefits from larger datasets rather than larger models.

\paragraph{Task-level analysis.}
On adult ECG interpretation tasks (CSN, CPSC2018, PTB-XL), SL models achieve moderate-to-high values of $\alpha$ ($0.24$--$0.8$) but consistently low values of $\beta$ ($<0.13$), suggesting that dataset size is the primary factor for improving supervised OOD generalization on standard diagnostic classification for supervised training strategies. 
For acute care predictions (mortality prediction and ICU admission prediction), a divergence emerges: Transformer-SL yields near-zero exponents ($\alpha = 0.028$ and $\beta = 0.011$ for mortality), indicating that scaling provides almost no OOD benefit in supervised settings, whereas Transformer-SSL and ResNet-SSL reach substantially larger exponents. 
On the structural heart disease detection and patient characteristics task (sex prediction), Transformer-SSL achieves notably higher values of $\beta$ on Echonext ($\beta = 0.366$) compared to Transformer-SL ($\beta = 0.004$). This reinforces the general finding that self-supervised paradigms are more effective at translating additional pre-training data into cross-domain transfer.

\begin{table*}[t]
\centering
\caption{OOD parameter scaling: best test loss per model size bin (lower is better, \textbf{bold} = best in bin). T-SL: Transformer-SL; T-SSL: Transformer-SSL; R-SL: ResNet-SL; R-SSL: ResNet-SSL.}
\label{tab:ood_param_scaling}
\setlength{\tabcolsep}{3.5pt}
\resizebox{\textwidth}{!}{%
\tiny
\begin{tabular}{l|cccc|cccc|cccc|cccc|cccc}
\toprule
\textbf{Dataset} & \multicolumn{4}{c}{$\mathbf{N \in [10^{4}, 10^{5})}$} & \multicolumn{4}{c}{$\mathbf{N \in [10^{5}, 10^{6})}$} & \multicolumn{4}{c}{$\mathbf{N \in [10^{6}, 10^{7})}$} & \multicolumn{4}{c}{$\mathbf{N \in [10^{7}, 10^{8})}$} & \multicolumn{4}{c}{$\mathbf{N \in [10^{8}, 10^{9})}$} \\
\cmidrule(lr){2-5}\cmidrule(lr){6-9}\cmidrule(lr){10-13}\cmidrule(lr){14-17}\cmidrule(lr){18-21}
 & T-SL & T-SSL & R-SL & R-SSL & T-SL & T-SSL & R-SL & R-SSL & T-SL & T-SSL & R-SL & R-SSL & T-SL & T-SSL & R-SL & R-SSL & T-SL & T-SSL & R-SL & R-SSL \\
\midrule
CPSC2018     & 0.3025 & 0.2918 & \textbf{0.1815} & 0.2848 & 0.2578 & 0.2524 & \textbf{0.1490} & 0.1999 & 0.2512 & 0.2350 & \textbf{0.1341} & 0.2016 & 0.2499 & 0.2117 & \textbf{0.1300} & 0.1859 & 0.2336 & 0.2068 & \textbf{0.1309} & 0.1722 \\
CSN          & 0.1313 & 0.1246 & \textbf{0.0869} & 0.1198 & 0.1224 & 0.1163 & \textbf{0.0744} & 0.0877 & 0.1156 & 0.1075 & \textbf{0.0704} & 0.0850 & 0.1145 & 0.0995 & \textbf{0.0684} & 0.0803 & 0.1119 & 0.0999 & \textbf{0.0683} & 0.0760 \\
Echonext     & 0.2364 & \textbf{0.2300} & 0.2412 & 0.2317 & 0.2264 & 0.2243 & 0.2321 & \textbf{0.2145} & 0.2263 & 0.2193 & 0.2295 & \textbf{0.2138} & 0.2256 & 0.2153 & 0.2284 & \textbf{0.2134} & 0.2205 & 0.2150 & 0.2284 & \textbf{0.2132} \\
MIMIC-ICU    & 0.2124 & 0.2069 & \textbf{0.2031} & 0.2079 & 0.2072 & 0.2008 & 0.2008 & \textbf{0.2001} & 0.2058 & \textbf{0.1988} & 0.2019 & 0.1998 & 0.2056 & \textbf{0.2001} & 0.2010 & 0.2003 & 0.2057 & \textbf{0.2001} & 0.2011 & 0.2004 \\
MIMIC-Mort.  & 0.2069 & 0.2106 & \textbf{0.2005} & 0.2088 & 0.2006 & 0.1969 & \textbf{0.1955} & 0.1960 & 0.1988 & 0.1936 & 0.1963 & \textbf{0.1927} & 0.1950 & \textbf{0.1893} & 0.1956 & 0.1932 & 0.1943 & \textbf{0.1878} & 0.1923 & 0.1918 \\
MIMIC-Sex    & 0.6642 & 0.6371 & 0.6321 & \textbf{0.5966} & 0.5982 & 0.6024 & 0.6179 & \textbf{0.5457} & 0.5624 & 0.5471 & 0.5957 & \textbf{0.5311} & 0.5607 & \textbf{0.5170} & 0.5993 & 0.5235 & 0.5608 & \textbf{0.5011} & 0.5766 & 0.5161 \\
PTB-XL(diag.)     & 0.0988 & 0.0918 & \textbf{0.0802} & 0.0830 & 0.0906 & 0.0861 & 0.0716 & \textbf{0.0681} & 0.0829 & 0.0810 & 0.0720 & \textbf{0.0698} & 0.0838 & 0.0782 & 0.0720 & \textbf{0.0689} & 0.0816 & 0.0770 & 0.0719 & \textbf{0.0669} \\
PTB-XL(sub.)    & 0.1688 & 0.1561 & \textbf{0.1330} & 0.1385 & 0.1500 & 0.1450 & 0.1184 & \textbf{0.1147} & 0.1389 & 0.1356 & 0.1190 & \textbf{0.1174} & 0.1401 & 0.1309 & \textbf{0.1155} & 0.1160 & 0.1362 & 0.1285 & 0.1161 & \textbf{0.1120} \\
PTB-XL(sup.)  & 0.5037 & 0.4555 & \textbf{0.3896} & 0.4016 & 0.4432 & 0.4311 & 0.3563 & \textbf{0.3415} & 0.4106 & 0.3978 & 0.3547 & \textbf{0.3440} & 0.4120 & 0.3798 & 0.3485 & \textbf{0.3400} & 0.3945 & 0.3776 & 0.3494 & \textbf{0.3269} \\
PTB-XL(form)     & 0.1944 & \textbf{0.1770} & 0.1917 & 0.1992 & 0.1840 & 0.1668 & 0.1774 & \textbf{0.1656} & 0.1814 & \textbf{0.1687} & 0.1695 & 0.1748 & 0.1876 & \textbf{0.1722} & 0.1800 & 0.1747 & 0.2054 & \textbf{0.1716} & 0.1927 & 0.1771 \\
PTB-XL(rhythm)     & 0.1039 & 0.0972 & \textbf{0.0739} & 0.1091 & 0.0882 & 0.0914 & \textbf{0.0581} & 0.0797 & 0.0863 & 0.0824 & \textbf{0.0531} & 0.0742 & 0.0837 & 0.0803 & \textbf{0.0496} & 0.0747 & 0.0805 & 0.0713 & \textbf{0.0499} & 0.0685 \\
\bottomrule
\end{tabular}
}
\end{table*}

In the parameter scaling regime (Table~\ref{tab:ood_param_scaling}), ResNet-based models demonstrate superior efficiency in small-to-medium ranges ($N < 10^6$), where ResNet-SL frequently achieves the lowest loss.
However, as model size grows ($N \ge 10^7$), the advantage shifts toward the SSL paradigm. 
Notably, Transformer-SSL consistently outperforms its SL counterpart across all bins and becomes the top performer in the largest parameter bin for datasets like MIMIC (sex) and PTB-XL (form), suggesting that Transformers require significant scale to overcome the ResNets' inherent  inductive biases.
Regarding data scaling (Table~\ref{tab:ood_data_scaling}), ResNet-SSL is robust in extremely low-data regimes ($D < 10^3$), winning the majority of datasets. 
As the pre-training dataset size scales toward $10^7$, ResNet-SL recovers dominance in several tasks (e.g., CPSC2018, PTB-XL (rhythm)), indicating that supervised learning may scale more effectively with larger dataset size for specific OOD tasks. 
While ResNets are more sample-efficient and robust at smaller scales, Transformers demonstrate stronger scaling potential when provided with sufficient parameters and data.

It should be noted that, in Figure~\ref{fig:param_data_scaling}, Transformer-SL models saturate and their ID performance decreases when the model size exceeds $10^7$, which is due to a data bottleneck and overfitting. This observation is consistent with previous studies~\citep{kaplan2020scaling}.

\begin{table*}[t]
\centering
\caption{OOD data scaling: best test loss per training set size bin (lower is better, \textbf{bold} = best in bin). T-SL: Transformer-SL; T-SSL: Transformer-SSL; R-SL: ResNet-SL; R-SSL: ResNet-SSL.}
\label{tab:ood_data_scaling}
\resizebox{\textwidth}{!}{%
\tiny
\setlength{\tabcolsep}{3pt}
\begin{tabular}{l|cccc|cccc|cccc|cccc|cccc|cccc}
\toprule
\textbf{Dataset} & \multicolumn{4}{c}{$\mathbf{D \in [10^{1}, 10^{2})}$} & \multicolumn{4}{c}{$\mathbf{D \in [10^{2}, 10^{3})}$} & \multicolumn{4}{c}{$\mathbf{D \in [10^{3}, 10^{4})}$} & \multicolumn{4}{c}{$\mathbf{D \in [10^{4}, 10^{5})}$} & \multicolumn{4}{c}{$\mathbf{D \in [10^{5}, 10^{6})}$} & \multicolumn{4}{c}{$\mathbf{D \in [10^{6}, 10^{7})}$} \\
\cmidrule(lr){2-5}\cmidrule(lr){6-9}\cmidrule(lr){10-13}\cmidrule(lr){14-17}\cmidrule(lr){18-21}\cmidrule(lr){22-25}
 & T-SL & T-SSL & R-SL & R-SSL & T-SL & T-SSL & R-SL & R-SSL & T-SL & T-SSL & R-SL & R-SSL & T-SL & T-SSL & R-SL & R-SSL & T-SL & T-SSL & R-SL & R-SSL & T-SL & T-SSL & R-SL & R-SSL\\
\midrule
CPSC2018    & 0.3285 & 0.3340 & 0.3305 & \textbf{0.2932} & 0.3300 & 0.2714 & 0.2963 & \textbf{0.2433} & 0.3040 & 0.2531 & 0.2360 & \textbf{0.2310} & 0.2777 & 0.2402 & \textbf{0.1777} & 0.2249 & 0.2675 & 0.2251 & \textbf{0.1375} & 0.2067 & 0.2499 & 0.2226 & \textbf{0.1276} & 0.1895 \\
CSN         & 0.1301 & 0.1366 & 0.1329 & \textbf{0.1145} & 0.1338 & 0.1220 & 0.1322 & \textbf{0.0939} & 0.1260 & 0.1173 & 0.1024 & \textbf{0.0867} & 0.1281 & 0.1092 & \textbf{0.0815} & 0.0823 & 0.1196 & 0.1016 & \textbf{0.0712} & 0.0839 & 0.1145 & 0.1068 & \textbf{0.0679} & 0.0811 \\
Echonext    & 0.2354 & 0.2389 & 0.2354 & \textbf{0.2267} & 0.2354 & 0.2241 & 0.2346 & \textbf{0.2197} & 0.2264 & 0.2244 & 0.2264 & \textbf{0.2186} & 0.2334 & 0.2211 & 0.2324 & \textbf{0.2182} & 0.2257 & 0.2167 & 0.2257 & \textbf{0.2151} & 0.2245 & 0.2177 & 0.2245 & \textbf{0.2128} \\
MIMIC-ICU   & 0.2194 & 0.2160 & 0.2157 & \textbf{0.2057} & 0.2181 & 0.2167 & 0.2247 & \textbf{0.2035} & 0.2151 & 0.2167 & 0.2083 & \textbf{0.2017} & 0.2156 & 0.2145 & 0.2045 & \textbf{0.2017} & 0.2130 & 0.2083 & 0.2011 & \textbf{0.2010} & 0.2127 & 0.2001 & 0.2010 & \textbf{0.1999} \\
MIMIC-Mort. & 0.2190 & \textbf{0.2077} & 0.2200 & 0.2116 & 0.2114 & \textbf{0.2040} & 0.2198 & 0.2058 & 0.2094 & \textbf{0.2001} & 0.2075 & 0.2021 & 0.2078 & 0.2015 & \textbf{0.1960} & 0.1961 & 0.1995 & 0.1932 & \textbf{0.1930} & 0.1945 & 0.1987 & \textbf{0.1893} & 0.1926 & 0.1915 \\
MIMIC-Sex   & 0.6425 & 0.6616 & 0.6856 & \textbf{0.6262} & 0.6395 & 0.6122 & 0.6879 & \textbf{0.5895} & 0.6235 & \textbf{0.5622} & 0.6373 & 0.5668 & 0.6262 & \textbf{0.5305} & 0.6372 & 0.5523 & 0.5706 & \textbf{0.5243} & 0.5928 & 0.5435 & 0.5623 & 0.5306 & 0.5878 & \textbf{0.5235} \\
PTB-XL(diag.)    & 0.0960 & 0.1019 & 0.1024 & \textbf{0.0851} & 0.0938 & 0.0862 & 0.0948 & \textbf{0.0738} & 0.0947 & 0.0834 & 0.0819 & \textbf{0.0736} & 0.0907 & 0.0822 & 0.0791 & \textbf{0.0733} & 0.0852 & 0.0796 & 0.0742 & \textbf{0.0711} & 0.0838 & 0.0801 & 0.0720 & \textbf{0.0689} \\
PTB-XL(sub.)   & 0.1668 & 0.1743 & 0.1700 & \textbf{0.1403} & 0.1626 & 0.1486 & 0.1594 & \textbf{0.1232} & 0.1602 & 0.1402 & 0.1366 & \textbf{0.1233} & 0.1534 & 0.1381 & 0.1299 & \textbf{0.1237} & 0.1429 & 0.1337 & 0.1222 & \textbf{0.1188} & 0.1401 & 0.1347 & 0.1184 & \textbf{0.1160} \\
PTB-XL(sup.) & 0.4908 & 0.5163 & 0.5195 & \textbf{0.4475} & 0.5014 & 0.4376 & 0.4841 & \textbf{0.3561} & 0.4741 & 0.4070 & 0.4022 & \textbf{0.3588} & 0.4535 & 0.4011 & 0.3842 & \textbf{0.3539} & 0.4195 & 0.3882 & 0.3599 & \textbf{0.3480} & 0.4120 & 0.3956 & 0.3506 & \textbf{0.3400} \\
PTB-XL(form)    & 0.2280 & 0.2063 & 0.2184 & \textbf{0.1895} & 0.2349 & 0.1909 & 0.2063 & \textbf{0.1726} & 0.2080 & 0.1801 & 0.1935 & \textbf{0.1706} & 0.1979 & \textbf{0.1731} & 0.1877 & 0.1768 & 0.1876 & \textbf{0.1694} & 0.1832 & 0.1742 & 0.1844 & \textbf{0.1701} & 0.1805 & 0.1730 \\
PTB-XL(rhythm)    & 0.1105 & 0.1119 & 0.1162 & \textbf{0.1001} & 0.1019 & 0.0977 & 0.1085 & \textbf{0.0867} & 0.1061 & 0.0901 & 0.0861 & \textbf{0.0779} & 0.1013 & 0.0831 & \textbf{0.0620} & 0.0742 & 0.0896 & 0.0791 & \textbf{0.0538} & 0.0765 & 0.0861 & 0.0810 & \textbf{0.0506} & 0.0730 \\
\bottomrule
\end{tabular}
}% end resizebox
\end{table*}

\subsection{Loss-to-loss scaling}
\label{appendix:loss-to-loss}
We provide the full parameter fits for the loss-to-loss scaling law (Eq.~\eqref{eq:loss_to_loss}) in Table~\ref{tab:loss2loss_joint}.

\begin{table*}[t]
\centering
\caption{Loss-to-loss scaling parameters fitted per architecture--paradigm and OOD dataset,
following Eq.~\eqref{eq:loss_to_loss} with $E_{\rm ID}$ and $E_{\rm OOD}$ fixed from joint scaling fits (Eq.~\eqref{eq:joint_scaling}).
Only $K$ and $\kappa$ are free parameters.
Task categories: (A) Adult ECG interpretation; (B) Cardiac structure \& function;
(C) Acute care prediction; (D) Patient characteristics.}
\label{tab:loss2loss_joint}
\resizebox{\textwidth}{!}{%
\scriptsize
\setlength{\tabcolsep}{4pt}
\begin{tabular}{cl|ccc|ccc|ccc|ccc}
\toprule
Cat. & Dataset
  & \multicolumn{3}{c}{Transformer-SL}
  & \multicolumn{3}{c}{ResNet-SL}
  & \multicolumn{3}{c}{Transformer-SSL}
  & \multicolumn{3}{c}{ResNet-SSL} \\
\cmidrule(lr){3-5}\cmidrule(lr){6-8}\cmidrule(lr){9-11}\cmidrule(lr){12-14}
& & $K$ & $\kappa$ & $R^2$
  & $K$ & $\kappa$ & $R^2$
  & $K$ & $\kappa$ & $R^2$
  & $K$ & $\kappa$ & $R^2$ \\
\midrule
\multirow{7}{*}{(A)}
& CPSC2018    & 0.486 & 0.174 & 0.86 & 0.448 & 0.200 & 0.94 & 0.081 & 0.433 & 0.92 & 0.125 & 0.224 & 0.88 \\
& CSN         & 0.163 & 0.088 & 0.84 & 0.164 & 0.141 & 0.93 & 0.025 & 0.443 & 0.88 & 0.026 & 0.376 & 0.81 \\
& PTB-XL(diagnostic)    & 0.114 & 0.075 & 0.81 & 0.065 & 0.164 & 0.95 & 0.011 & 0.495 & 0.83 & 0.080 & 0.050 & 0.84 \\
& PTB-XL(subclass)   & 0.204 & 0.094 & 0.83 & 0.133 & 0.118 & 0.92 & 0.018 & 0.558 & 0.85 & 0.134 & 0.050 & 0.82 \\
& PTB-XL(superclass) & 0.624 & 0.105 & 0.82 & 0.336 & 0.199 & 0.94 & 0.059 & 0.535 & 0.84 & 0.048 & 0.301 & 0.83 \\
& PTB-XL(form)    & 0.239 & 0.480 & 0.89 & 0.086 & 0.138 & 0.84 & 0.002 & 1.628 & 0.81 & 0.013 & 0.060 & 0.63 \\
& PTB-XL(rhythm)    & 0.159 & 0.162 & 0.88 & 0.152 & 0.174 & 0.93 & 0.061 & 0.209 & 0.85 & 0.024 & 0.341 & 0.80 \\
\midrule
\multirow{1}{*}{(B)}
& Echonext    & 0.265 & 0.050 & 0.78 & 0.015 & 0.051 & 0.82 & 0.020 & 0.339 & 0.86 & 0.013 & 0.347 & 0.83 \\
\midrule
\multirow{2}{*}{(C)}
& MIMIC(icu)   & 0.249 & 0.050 & 0.92 & 0.188 & 0.050 & 0.86 & 0.204 & 0.050 & 0.84 & 0.005 & 0.401 & 0.81 \\
& MIMIC(mortality) & 0.244 & 0.057 & 0.79 & 0.068 & 0.175 & 0.91 & 0.193 & 0.050 & 0.85 & 0.018 & 0.438 & 0.91 \\
\midrule
\multirow{1}{*}{(D)}
& MIMIC(sex)   & 0.768 & 0.075 & 0.72 & 0.802 & 0.050 & 0.75 & 0.130 & 0.384 & 0.87 & 0.079 & 0.383 & 0.94 \\
\midrule
\multicolumn{2}{l|}{\textit{Mean}}
  & 0.320 & 0.128 & 0.83 & 0.223 & 0.133 & 0.89 & 0.073 & 0.466 & 0.85 & 0.051 & 0.270 & 0.83 \\
\multicolumn{2}{l|}{\textit{Std}}
  & 0.202 & 0.118 & 0.05 & 0.219 & 0.056 & 0.06 & 0.069 & 0.405 & 0.03 & 0.044 & 0.143 & 0.08 \\
\bottomrule
\end{tabular}
}
\end{table*}

\subsection{Analysis of OOD absolute performance}
\label{sec:supplementary_crossover}

\begin{figure}[t]
    \centering
    \includegraphics[width=1.0\linewidth]{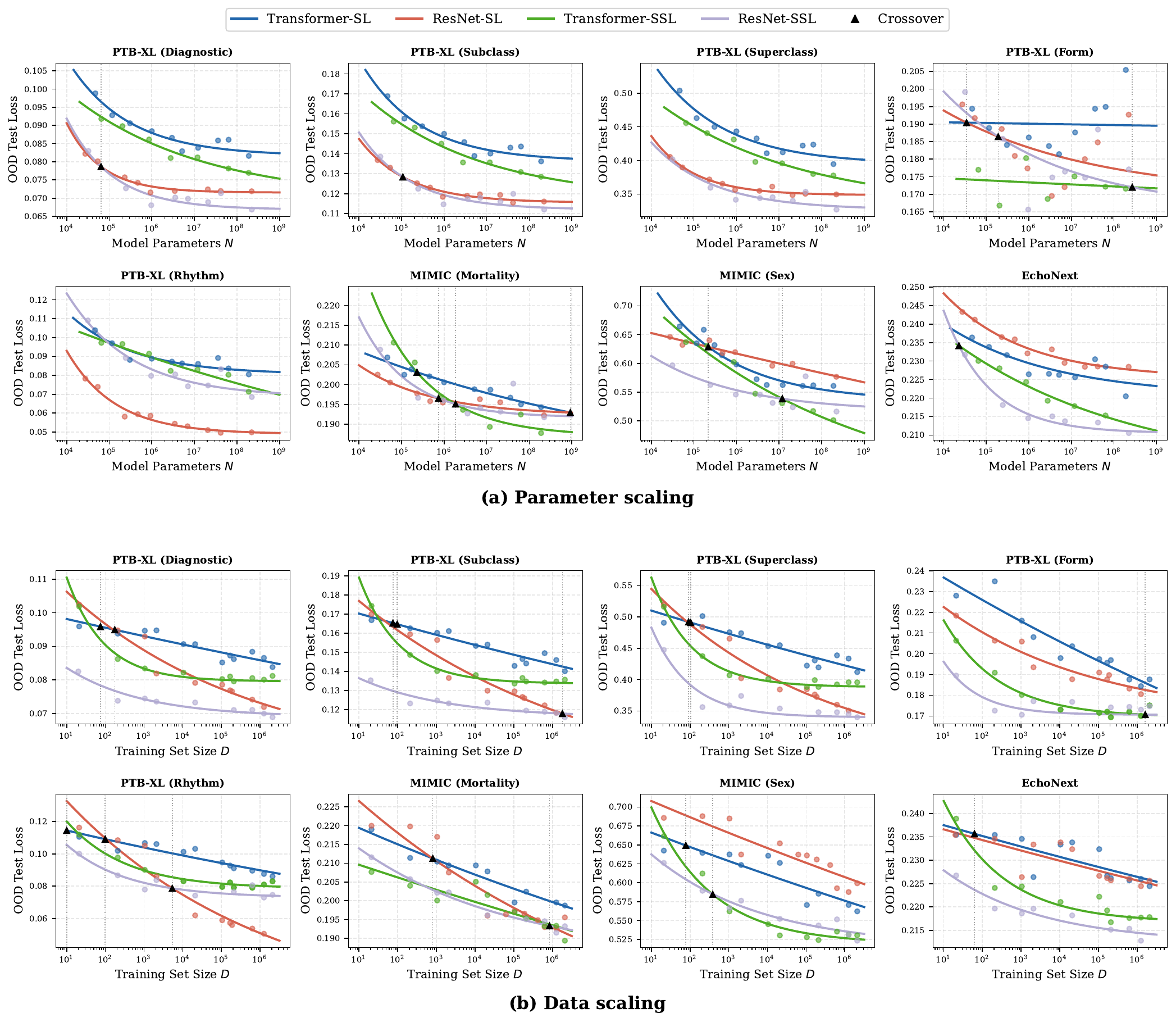}
    \vspace{-2mm}
    \caption{
    OOD scaling behaviors and paradigm crossovers across downstream tasks.
    Solid lines represent fitted scaling curves, while black triangles denote crossover points. 
    (a) Parameter scaling (pre-trained on 90\% CODE): ResNet-based models generally attain the lowest absolute OOD loss, confirming a strong architectural inductive bias. For Transformers, SSL consistently outperforms SL across all observed scales. 
    (b) Data scaling ($\sim 10^7$ parameters): A distinct paradigm crossover emerges in most datasets, with SSL models dominating the low-data regime. Notably, SSL consistently yields superior performance over SL on unseen clinical tasks.}
    \label{fig:ood_absolute_loss}
\end{figure}

In parameter scaling (Figure~\ref{fig:ood_absolute_loss}(a)), ResNet-SL attains the lowest absolute OOD loss on the majority of datasets, confirming the strong inductive bias of convolutional architectures. Across all datasets, Transformers exhibit a consistent paradigm gap: Transformer-SSL models substantially outperform Transformer-SL models at every observed scale. 
In data scaling (Figure~\ref{fig:ood_absolute_loss}(b)), a paradigm crossover is visible on datasets: SSL models achieve lower loss in the low-data regime, but SL models overtake them as the pre-training dataset size increases.
On tasks distant from the pre-training distribution, like  sex prediction and structural heart disease detection, SSL models maintain lower loss over SL models, suggesting that the benefit of self-supervised representations is most pronounced when the downstream task diverges from the source labels. 
Transformer-SSL overtakes ResNet-SSL at very large model sizes, as opposed to large data sizes. 
This finding is supported by our benchmarking results (Appendix~\ref{appendix:benchmark}), which show that Transformer-SSL models excel at very large capacity.

\subsection{Analysis of compute-optimal allocation}
\label{sec:appendix_compute}
\begin{figure}[t]
    \centering
    \includegraphics[width=1.0\linewidth]{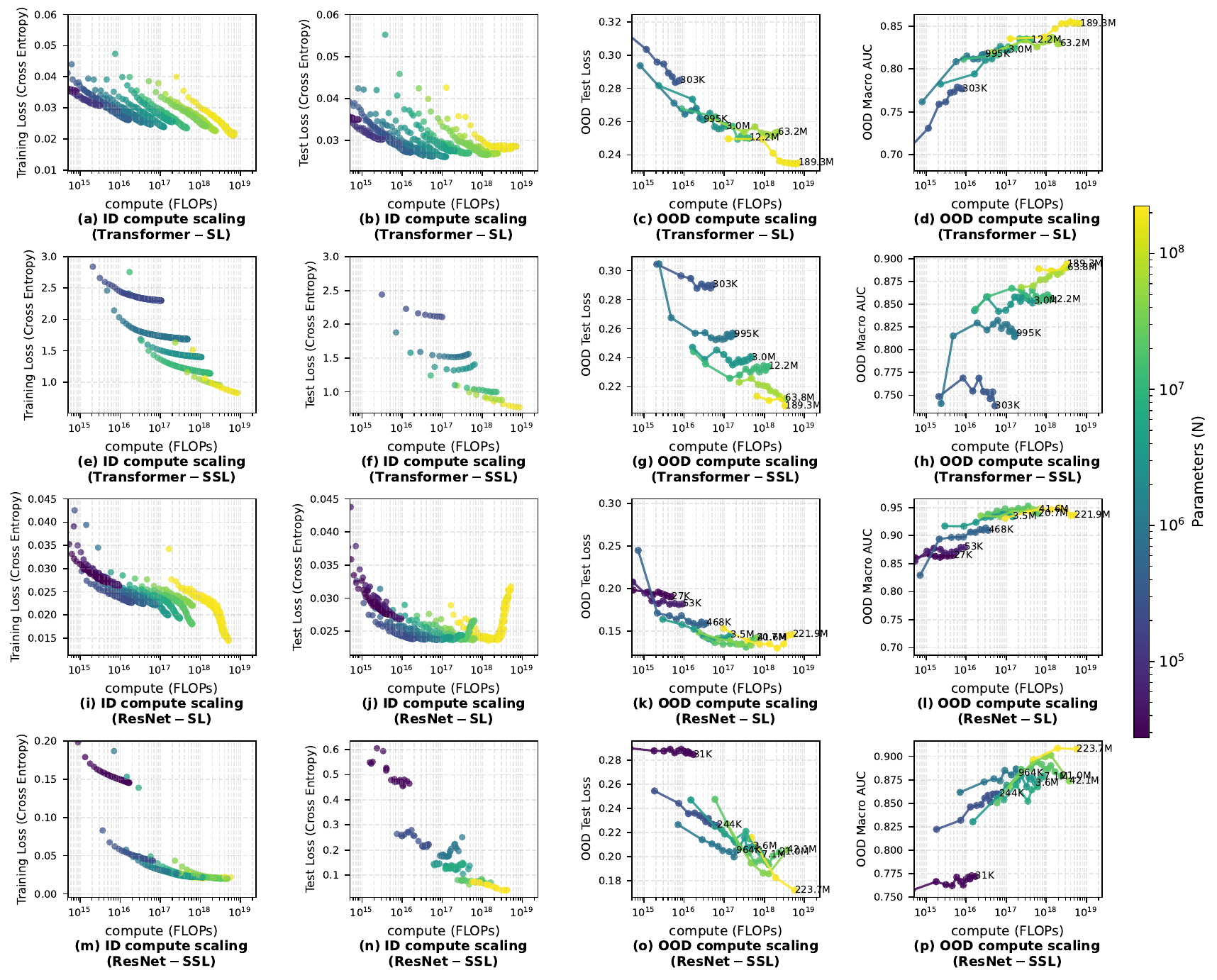}
    \caption{We show how ID training loss, ID test loss, OOD test loss, and OOD macro-AUC change with increasing total training compute. The pre-training dataset consists of 90\% of CODE, and ID performance is evaluated on the remaining 10\% held-out test set, whereas OOD evaluation is conducted on the CPSC2018 dataset. All evaluations are performed using linear probing.}
    \label{fig:compute_scaling}
\end{figure}

\vspace{-2mm}
As illustrated in Figure~\ref{fig:compute_scaling}, supervised and self-supervised pre-training exhibit distinct ID and OOD scaling trajectories under varying computational budgets. 
The frontiers in Figure~\ref{fig:compute_scaling_frontier} are defined as $L^*(C) = a(C+c)^{-b} + d$, where $b$ determines the compute scaling efficiency, while $d$ represents the irreducible loss limit as $C \to \infty$ due to task-inherent difficulty. The offset $c$ accounts for the saturation at the low-compute regime. 
Our analysis reveals distinct paradigm biases. For Transformers, SSL significantly steepens the slope from $b=0.133$ (Transformer-SL) to $0.184$ (Transformer-SSL), while lowering the theoretical floor from $d=0.209$ (Transformer-SL) to $0.184$ (Transformer-SSL), proving SSL's superior OOD generalization at large scale. Interestingly, ResNet-SL exhibits the most aggressive scaling ($b=0.358$) on this specific task, highlighting the architectural inductive biases.

To determine the optimal allocation of compute between data size and model size for each architecture-paradigm combination, we derive compute-optimal allocations over varying model and dataset sizes, and analytically minimize the loss under our modified compute constraint.

The compute-optimal allocation framework of \citet{hoffmann2022training} was developed for large language models trained for a single epoch, where the number of training tokens equals the number of unique tokens the model encounters and the total compute is simply $C = 6ND$. In that setting there is no distinction between unique data and total processed data. Our experimental setting differs in two respects. 
First, instead of a single pass, models are trained for multiple epochs until they reach the maximum number of steps, where all models empirically converge. Consequently, the same sample is processed many times.
Second, the variable $D$ in our scaling law (Eq.~\eqref{eq:joint_scaling}) refers to the number of unique training samples rather than the seen samples. These differences require us to revisit the relationship between total compute and the scaling-law variables before deriving optimal allocations.
 
Let $\mathrm{MACs}_{\mathrm{fwd}}(N)$ denote the forward-pass multiply-accumulate operations per sample for a model of size $N$, and let $E_{\mathrm{ep}}(N,D)$ denote the number of epochs required to reach convergence. The total training compute is then $C = 6 \cdot \mathrm{MACs}_{\mathrm{fwd}}(N) \cdot D \cdot E_{\mathrm{ep}}(N,D)$. In principle $E_{\mathrm{ep}}$ could depend on both $N$ and $D$, but in our experiments we observe that the epoch count at convergence is approximately stable across configurations. We therefore assume that $E_{\mathrm{ep}}(N,D) \approx \bar{E}_{\mathrm{ep}}$ is a configuration-independent constant. Combined with the standard approximation $\mathrm{MACs}_{\mathrm{fwd}}(N) \propto~N$ \citep{hoffmann2022training}, the compute constraint simplifies to $C = \kappa \, N \,D$, where $\kappa$ absorbs all proportionality constants. This has the same functional form as the single-epoch case.
 
Given the joint scaling law $L(N,D) = E + A \cdot N^{-\alpha} + B \cdot D^{-\beta}$
and the constraint $C = \kappa ND$, we minimize $L$ for a fixed compute budget $C$ via the Lagrangian
\begin{equation}
    \mathcal{L} = E + A\,N^{-\alpha} + B\,D^{-\beta}
                  - \lambda\bigl(\kappa\,N\,D - C\bigr).
\end{equation}

The first-order conditions are:
\begin{align}
    \frac{\partial \mathcal{L}}{\partial N} = 0
    \;\;&\Longrightarrow\;\;
    \alpha\,A\,N^{-\alpha-1} = \lambda\,\kappa\,D,
    \label{eq:foc_N} \\[4pt]
    \frac{\partial \mathcal{L}}{\partial D} = 0
    \;\;&\Longrightarrow\;\;
    \beta\,B\,D^{-\beta-1} = \lambda\,\kappa\,N.
    \label{eq:foc_D}
\end{align}
Dividing Eq.~\eqref{eq:foc_N} by Eq.~\eqref{eq:foc_D} eliminates $\lambda$
and yields the balance condition:
\begin{equation}
    \alpha \cdot A\,N^{-\alpha}
    \;=\;
    \beta \cdot B\,D^{-\beta},
    \label{eq:balance}
\end{equation}
stating that the marginal loss reductions from scaling model size and data are
equalized at the optimum. Solving Eq.~\eqref{eq:balance} for $D$ and
substituting into the constraint $C = \kappa ND$ gives:
\begin{equation}
    N^*(C) \;\propto\; C^{\,\beta/(\alpha+\beta)},
    \qquad
    D^*(C) \;\propto\; C^{\,\alpha/(\alpha+\beta)}.
    \label{eq:ND_opt}
\end{equation}
Plugging these back into the scaling law yields the optimal loss
$L^*(C) = E + G \cdot C^{-\gamma}$, where $\gamma = \alpha\beta/(\alpha+\beta)$ is the compute efficiency, the rate at which loss decreases per unit increase in compute under optimal allocation.
Crucially, $\bar{E}_{\mathrm{ep}}$ is absorbed into $\kappa$ and affects only the prefactors of $N^*$ and $D^*$, not the exponents: it shifts the optimal frontier vertically on a log--log plot without changing its slope. The exponents themselves reflect a simple trade-off: when $\alpha \gg \beta$, model scaling saturates quickly and most additional compute goes to data; when $\beta \gg \alpha$, the reverse holds; when $\alpha \approx \beta$, compute is split roughly equally.
For example, if $\alpha = 0.4$ and $\beta = 0.1$, then $N^* \propto C^{0.2}$ and $D^* \propto C^{0.8}$: under optimal allocation, 80\% of a marginal log-increase in compute should be directed toward enlarging the dataset and only 20\% toward growing the model.

If the constant-epoch assumption does not hold and the epoch count instead follows a power law $E_{\mathrm{ep}} \propto N^{\gamma_N} D^{\gamma_D}$, the compute constraint becomes $C \propto N^{a}\,D^{b}$ with $a = 1+\gamma_N$ and $b = 1+\gamma_D$. Repeating the Lagrangian analysis yields the generalized allocations $N^* \propto C^{\,\beta/(a\beta+\alpha b)}$ and $D^* \propto C^{\,\alpha/(a\beta+\alpha b)}$, which reduce to Eq.~\eqref{eq:ND_opt} when $a = b = 1$.
 
\begin{figure}[H]
    \centering
    \includegraphics[width=1.0\linewidth]{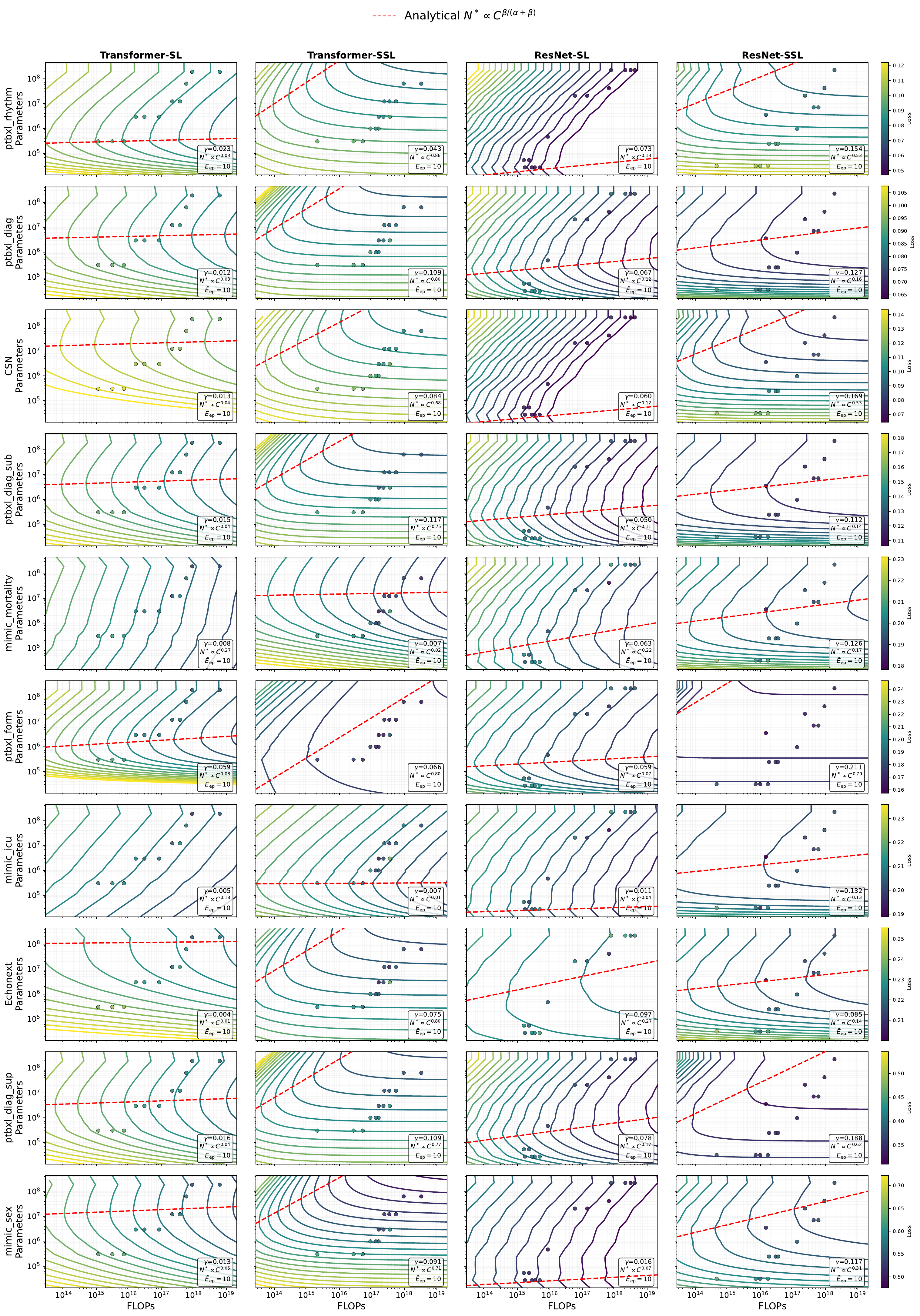}
    \caption{Compute-optimal allocation across 10 OOD datasets
and 4 model configurations. Loss contours are derived from the fitted scaling law
$L(N,D) = E + AN^{-\alpha} + BD^{-\beta}$ (Eq.~\eqref{eq:joint_scaling}). The red dashed line shows the closed-form compute-optimal model size $N^* \propto C^{\beta/(\alpha+\beta)}$, obtained by Lagrangian optimization. Each panel is annotated with the compute efficiency $\gamma = \alpha\beta/(\alpha+\beta)$ and the optimal scaling exponent.
Under compute-optimal allocation, ResNet-SL and Transformer-SL direct more than $70\%$ of marginal compute to data, while Transformer-SSL favors larger models over more data.
}
    \label{fig:optimal_allocation_10x4}
\end{figure}

The iso-loss contour plots in Figure~\ref{fig:compute_scaling_frontier} and Figure~\ref{fig:optimal_allocation_10x4} visualize these relationships in the $(C, N)$ plane. For each grid point $(C, N)$, we compute the unique dataset size that the compute budget can support via $D_{\mathrm{unique}} = C / (6\cdot\mathrm{MACs}_{\mathrm{fwd}}(N) \cdot \bar{E}_{\mathrm{ep}})$ and evaluate the converged test loss $L(N, D_{\mathrm{unique}})$. Each contour line thus connects $(C, N)$ pairs that yield the same predicted loss when the full budget is spent training to convergence. 
We observe a clear paradigm bias, with supervised models consistently favoring more data over larger models. In contrast, Transformer-SSL consistently exhibits a steeper slope, indicating a preference for larger models to achieve better downstream transfer. 
For specific downstream tasks, such as sex and mortality prediction, the optimal model size of Transformer-SL is smaller than $10^5$ parameters. 
This suggests that deploying smaller models trained on larger datasets is favorable for transfer to these unseen tasks.

\subsection{Analysis of label scaling}
\label{appendix:label_scaling}
\begin{figure}[t]
    \centering
    \includegraphics[width=0.9\linewidth]{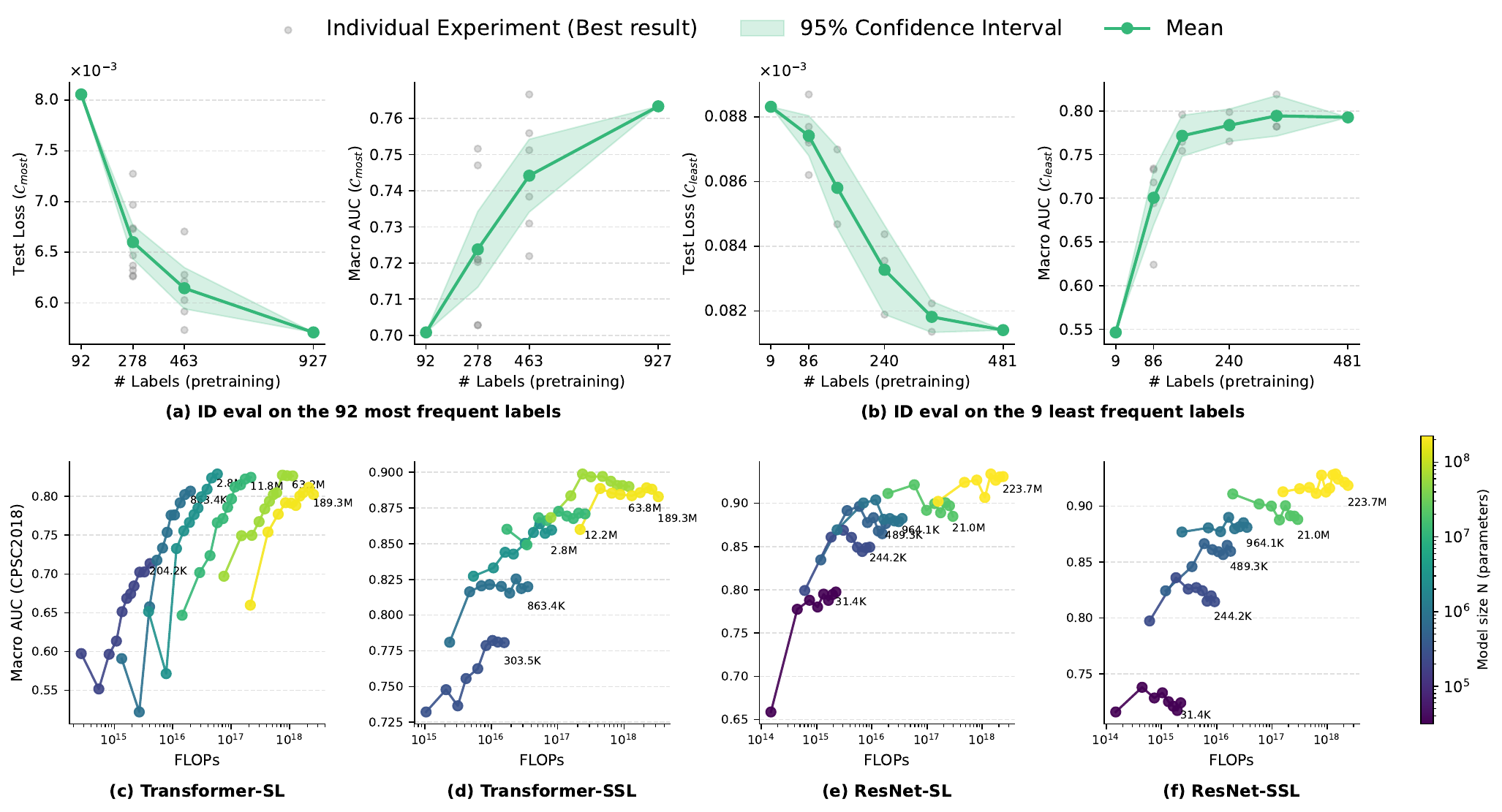}
    \caption{Extended results for Figure~\ref{fig:label_scaling}. In multi-label, long-tailed learning regimes, Transformer-SL exhibits parameter scaling bottlenecks, significantly underperforming its SSL counterpart. Conversely, ResNet-SL maintains robust scaling at larger model size, outperforming ResNet-SSL.}
    \label{fig:id_label_scaling_1x4}
\end{figure}

As shown in Figure~\ref{fig:id_label_scaling_1x4}, the macro-AUROC trends align with the test loss observed in Figure~\ref{fig:label_scaling}. Due to the inherent challenges of multi-label learning, self-supervised learning exhibits a more robust parameter scaling trajectory compared to supervised pre-training, particularly for Transformers. 
This suggests that the multi-label setting is challenging for supervised pre-training, and that ResNet models exhibit more stable model size scaling than Transformers.

\subsection{Quantifying overfitting from finite data}

\begin{figure}[t]
    \centering
    \includegraphics[width=1.0\linewidth]{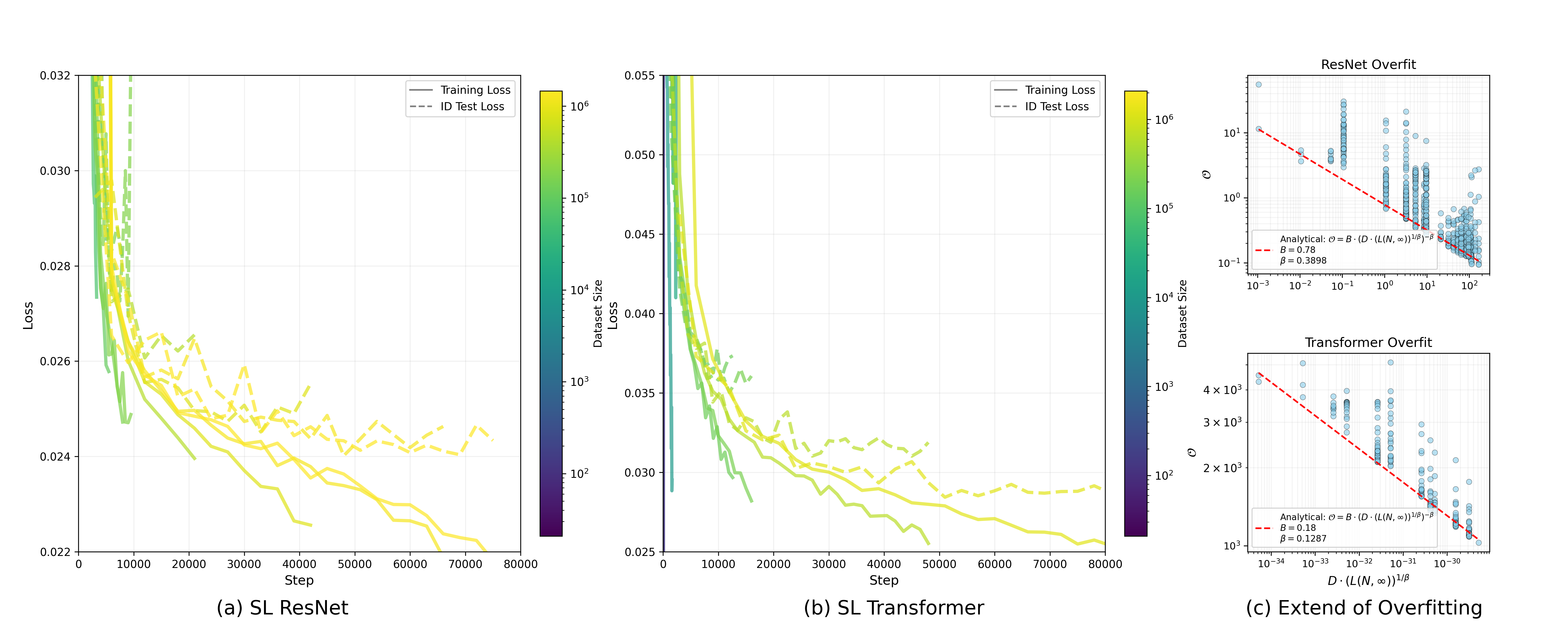}
    \caption{ 
    Panels (a) and (b) present the training and validation cross-entropy loss trajectories on the CODE dataset for a fixed model size of $N=10^7$.
    Panel (c) illustrates the verification of the \textit{extent of overfitting}, defined as $\mathcal{O} = (L(N, D) - L(N, \infty)) / L(N, \infty)$. We plot $\mathcal{O}$ against the characteristic variable $X = D \cdot (L(N, \infty))^{1/\beta}$. 
    The experimental results (blue scatters) align closely with the analytical power-law curves (red dashed lines) $\mathcal{O} = B \cdot X^{-\beta}$.}
    \label{fig:extend_overfitting}
\end{figure}

\begin{figure}[t]
    \centering
    \includegraphics[width=0.6\linewidth]{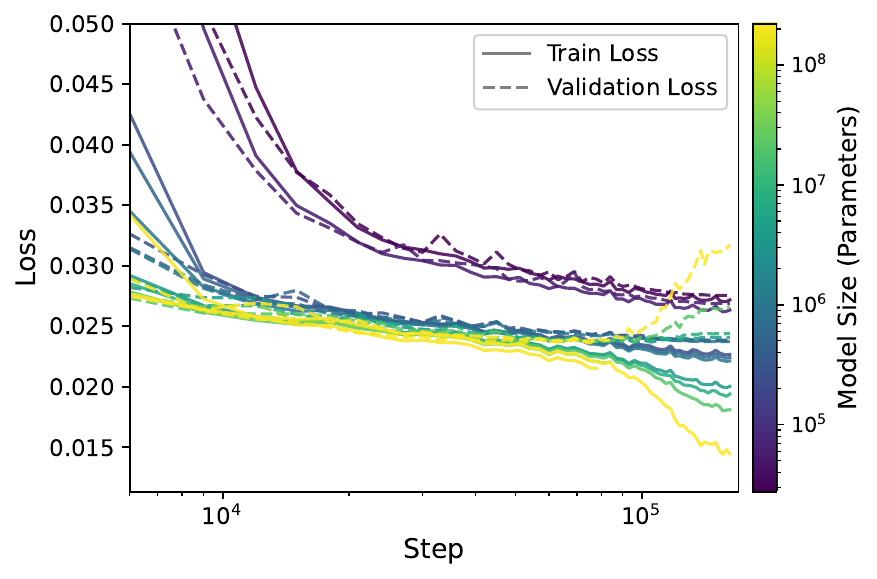}
    \caption{ 
    The training and test loss trajectories of ResNet-SL. Larger models exhibit faster convergence compared to their smaller counterparts in terms of training steps.}
    \label{fig:model_overfit}
\end{figure}

As shown in Figure~\ref{fig:extend_overfitting} and Figure~\ref{fig:model_overfit}, the loss trajectories for SL models on the CODE dataset are reported. Empirically, reducing the training set size substantially widens the gap between training and
validation loss.
To quantify this effect, we define the \emph{extent of overfitting} as the relative deviation of the observed validation loss from the irreducible floor. This floor, $L(N,\infty) = E + A\,N^{-\alpha}$
(see~\eqref{eq:marginal_scaling}), represents the minimum achievable loss for a model of size $N$ in the limit of infinite data. Accordingly, the extent of overfitting is
\begin{equation}
  \mathcal{O}(N, D)
  = \frac{L(N, D) - L(N, \infty)}{L(N, \infty)}
  = \frac{B\,D^{-\beta}}{L(N, \infty)}.
\end{equation}
The collapse of all configurations onto a single curve in panel~(c) enables a direct prediction of the dataset size required to maintain a prescribed overfitting budget. Setting a tolerance $\mathcal{O} \le
\epsilon$ (e.g., $\epsilon = 0.1$, corresponding to a 10\% excess loss due to finite data) yields the minimum data requirement $D_{\min} \propto \bigl(L(N,\infty)\bigr)^{-1/\beta}$.
Because $\beta$ is small (particularly for Transformers), this relationship is strongly non-linear: halving the irreducible loss requires a multiplicative increase in dataset size far exceeding a factor of two.

\subsection{Benchmarking ECG foundation models}
\label{appendix:benchmark}
\paragraph{Baselines.}
We consider two major families of model architectures: CNN-based and Transformer-based.
Within each family, we include both models trained from scratch on downstream tasks and foundation models fine-tuned from pre-trained checkpoints.
 
For CNN-based models trained from scratch, we evaluate
XResNet1d101~\citep{he2019bag},
ResNet (Wang)~\citep{wang2017time},
InceptionTime~\citep{ismail2020inceptiontime},
TimesNet~\citep{wu2022timesnet},
TCN~\citep{lea2016temporal},
MICN~\citep{wang2023micn},
and SCINet~\citep{liu2022scinet}.
For Transformer-based models trained from scratch, we evaluate
Autoformer~\citep{wu2021autoformer},
Informer~\citep{zhou2021informer},
Reformer~\citep{kitaev2020reformer},
FEDformer~\citep{zhou2022fedformer},
Crossformer~\citep{zhang2023crossformer},
PatchTST~\citep{nie2022time},
and iTransformer~\citep{liu2023itransformer}.
 
For foundation models, we organize baselines into three categories based on architecture and pre-training paradigm.
ResNet-SL models include ResNet (Ribeiro)~\citep{ribeiro2020automatic}, a 7.1M-parameter model pre-trained with supervised labels on 2.3M ECGs, and ECGFounder~\citep{li2024electrocardiogram}, the largest CNN baseline with 30.7M parameters pre-trained on 10M recordings.
ResNet-SSL models include ResNet (PCLR)~\citep{diamant2022patient}, which uses patient-level contrastive learning on 0.4M samples, and ResNet (Merl)~\citep{liu2024zero}, pre-trained via multi-modal contrastive learning on 0.8M ECG--report pairs.
Transformer-SSL models include ViT (Merl)~\citep{liu2024zero}, the Transformer counterpart of ResNet (Merl) trained under the same SSL framework; HuBERT (Base)~\citep{coppola2024hubert}, a 93M-parameter BERT-style model adapted from speech and pre-trained on 9.1M ECGs; ECG-FM~\citep{mckeen2024ecg}, a 90.9M-parameter model pre-trained with masked auto-encoding on 1.5M samples through a hybrid contrastive and generative self-supervised learning framework; and HeartLang~\citep{jin2025reading}, which employs a heartbeat-level tokenizer and is pre-trained on 0.8M recordings.
We note that no Transformer-SL foundation model is currently available as a public baseline.

\begin{table}[t]
\caption{Fine-tuning configurations for foundation models. All models use the AdamW optimizer. Learning rates and strategies follow the original papers.}
\label{tab:ft_config}
\centering
\small
\setlength{\tabcolsep}{4pt}
\renewcommand{\arraystretch}{1.15}
\begin{tabular}{l|c|cc|l}
\toprule
\multirow{2}{*}{Model} &
\multirow{2}{*}{Learning Rate} &
\multicolumn{2}{c|}{Input Configuration} &
\multirow{2}{*}{Fine-tuning Strategy} \\
& & Freq.\ (Hz) & Length & \\
\midrule
ResNet (PCLR)    & 1e-3 & 500 & 4{,}096 & Full-tuning \\
ResNet (Ribeiro) & 1e-3 & 400 & 4{,}096 & Full-tuning \\
ResNet (Merl)    & 1e-3 & 500 & N/A     & Full-tuning \\
ECGFounder       & 1e-4 & 500 & 5{,}000 & Full-tuning \\
\midrule
ViT (Merl)       & 1e-3 & 500 & N/A     & Full-tuning \\
ECG-FM           & 1e-6 & 500 & N/A & Full-tuning \\
HeartLang        & 1e-3 & 100 & N/A     & Full-tuning \\
HuBERT (Base)    & 1e-5 & 500 & 2{,}500 & \makecell[l]{Full-tuning except \\ conv.\ embedder (frozen)} \\
\bottomrule
\end{tabular}
\end{table}

\begin{table*}[b!]
\caption{Full macro-AUROC results across all benchmarks with model specifications.
PTB-XL results are reported as the average macro-AUROC across five subsets, with the full results provided in Table~\ref{tab:ptbxl_results_only}.
Mean is computed over all six benchmark columns.
Best and second-best are computed within each architecture group.
Pre-trained models are highlighted with a blue background.}
\label{tab:benchmark}
\centering
\scriptsize
\setlength{\tabcolsep}{2.5pt}
\renewcommand{\arraystretch}{1.2}
\begin{tabular}{l|cccccc|c|c|c|c}
\toprule
\multirow{2}{*}{Method} &
PTB-XL & EchoNext & MIMIC & MIMIC & MIMIC & CPSC &
\multirow{2}{*}{Mean} &
\multirow{2}{*}{\#Param} &
\multirow{2}{*}{\#Data} &
\multirow{2}{*}{Paradigm} \\
& (avg) & & (Sex) & (ICU) & (Mort.) & 2018 & & & & \\
\midrule
\multicolumn{11}{c}{\textbf{CNN}} \\
XResNet1d101~\citep{he2019bag}
  & 0.904 & 0.794 & 0.751 & 0.662 & 0.705 & 0.961
  & 0.796 & 1.9M & / & SL \\
ResNet (Wang)~\citep{wang2017time}
  & \second{0.919} & 0.780 & 0.750 & 0.666 & 0.711 & 0.959
  & 0.797 & 451K & / & SL \\
InceptionTime~\citep{ismail2020inceptiontime}
  & 0.915 & 0.791 & 0.778 & 0.690 & 0.737 & \second{0.962}
  & 0.812 & 513K & / & SL \\
TimesNet~\citep{wu2022timesnet}
  & 0.783 & 0.734 & 0.748 & 0.660 & 0.702 & 0.806
  & 0.739 & 61.9M & / & SL \\
TCN~\citep{lea2016temporal}
  & 0.820 & 0.760 & 0.769 & 0.682 & 0.729 & 0.842
  & 0.767 & 11.7M & / & SL \\
MICN~\citep{wang2023micn}
  & 0.754 & 0.705 & 0.766 & 0.570 & 0.612 & 0.797
  & 0.701 & 6.3M & / & SL \\
SCINet~\citep{liu2022scinet}
  & 0.743 & 0.680 & 0.771 & 0.588 & 0.626 & 0.788
  & 0.699 & 616K & / & SL \\
\cellcolor{ftblue}ResNet (PCLR)~\citep{diamant2022patient}
  & \ftcell{0.860} & \ftcell{0.800}
  & \ftcell{\best{0.927}} & \ftcell{0.729}
  & \ftcell{\best{0.838}} & \ftcell{0.879}
  & \ftcell{0.839} & \ftcell{6M} & \ftcell{0.4M} & \ftcell{SSL} \\
\cellcolor{ftblue}ResNet (Ribeiro)~\citep{ribeiro2020automatic}
  & \ftcell{0.906} & \ftcell{\second{0.800}}
  & \ftcell{0.913} & \ftcell{\second{0.755}}
  & \ftcell{0.816} & \ftcell{0.948}
  & \ftcell{\second{0.856}} & \ftcell{7.1M} & \ftcell{2.3M} & \ftcell{SL} \\
\cellcolor{ftblue}ResNet (Merl)~\citep{liu2024zero}
  & \ftcell{0.916} & \ftcell{\best{0.807}}
  & \ftcell{0.890} & \ftcell{0.720}
  & \ftcell{0.810} & \ftcell{0.945}
  & \ftcell{0.848} & \ftcell{3.9M} & \ftcell{0.8M} & \ftcell{SSL} \\
\cellcolor{ftblue}ECGFounder~\citep{li2024electrocardiogram}
  & \ftcell{\best{0.925}} & \ftcell{0.793}
  & \ftcell{\second{0.915}} & \ftcell{\best{0.762}}
  & \ftcell{\second{0.832}} & \ftcell{\best{0.968}}
  & \ftcell{\best{0.866}} & \ftcell{30.7M} & \ftcell{10M} & \ftcell{SL} \\
\midrule
\multicolumn{11}{c}{\textbf{Transformer}} \\
Autoformer~\citep{wu2021autoformer}
  & 0.711 & 0.705 & 0.723 & 0.637 & 0.675 & 0.822
  & 0.712 & 6.0M & / & SL \\
Informer~\citep{zhou2021informer}
  & 0.741 & 0.753 & 0.819 & 0.697 & 0.750 & 0.797
  & 0.760 & 5.6M & / & SL \\
Reformer~\citep{kitaev2020reformer}
  & 0.756 & 0.731 & 0.808 & 0.680 & 0.737 & 0.810
  & 0.754 & 6.0M & / & SL \\
FEDformer~\citep{zhou2022fedformer}
  & 0.734 & 0.703 & 0.770 & 0.668 & 0.710 & 0.814
  & 0.733 & 6.6M & / & SL \\
Crossformer~\citep{zhang2023crossformer}
  & 0.833 & 0.740 & 0.779 & 0.687 & 0.575 & 0.876
  & 0.748 & 4.0M & / & SL \\
PatchTST~\citep{nie2022time}
  & 0.804 & 0.728 & 0.823 & 0.706 & 0.773 & 0.868
  & 0.784 & 17.4M & / & SL \\
iTransformer~\citep{liu2023itransformer}
  & 0.711 & 0.723 & 0.763 & 0.665 & 0.707 & 0.705
  & 0.712 & 593K & / & SL \\
\cellcolor{ftblue}ViT (Merl)~\citep{liu2024zero}
  & \ftcell{0.883} & \ftcell{0.751}
  & \ftcell{0.817} & \ftcell{0.684}
  & \ftcell{0.759} & \ftcell{0.934}
  & \ftcell{0.805} & \ftcell{5.5M} & \ftcell{0.8M} & \ftcell{SSL} \\
\cellcolor{ftblue}HuBERT (Base)~\citep{coppola2024hubert}
  & \ftcell{0.892} & \ftcell{\second{0.758}}
  & \ftcell{\second{0.885}} & \ftcell{\second{0.750}}
  & \ftcell{\second{0.814}} & \ftcell{0.936}
  & \ftcell{\second{0.839}} & \ftcell{93M} & \ftcell{9.1M} & \ftcell{SSL} \\
\cellcolor{ftblue}ECG-FM~\citep{mckeen2024ecg}
  & \ftcell{\best{0.913}} & \ftcell{\best{0.819}}
  & \ftcell{\best{0.941}} & \ftcell{\best{0.789}}
  & \ftcell{\best{0.863}} & \ftcell{\best{0.969}}
  & \ftcell{\best{0.882}} & \ftcell{90.9M} & \ftcell{1.5M} & \ftcell{SSL} \\
\cellcolor{ftblue}HeartLang~\citep{jin2025reading}
  & \ftcell{\second{0.906}} & \ftcell{0.755}
  & \ftcell{0.876} & \ftcell{0.736}
  & \ftcell{0.799} & \ftcell{\second{0.947}}
  & \ftcell{0.837} & \ftcell{39.1M} & \ftcell{0.8M} & \ftcell{SSL} \\
\bottomrule
\end{tabular}
\end{table*}

\paragraph{Experimental Setup.}
We evaluate all foundation models using fine-tuning rather than linear probing to ensure each model achieves its optimal downstream performance through hyperparameter optimization. Given that different foundation models utilize distinct training protocols, we adhere to the best practices specified in their original papers, incorporating minor adjustments tailored to each dataset. 
In practice, we employ full-parameter fine-tuning for most models; the notable exception is HuBERT, for which we freeze the convolutional feature extractor as recommended by the authors. Furthermore, each foundation model’s input signal is resampled and converted to a tensor, then either truncated or padded to match the specific sequence length required by its respective pre-training configuration.
Detailed hyperparameters, input configurations, and fine-tuning strategies for each model are summarized in Table~\ref{tab:ft_config}.
For XResNet1d101, ResNet (Wang), and InceptionTime, we use the same hyperparameter settings as in previous benchmarks~\citep{strodthoff2020deep}.
For the remaining from-scratch baselines, we follow the hyperparameter recommended by \cite{wang2024tssurvey}.

\paragraph{Result analysis.}
We report the full benchmark results in Table~\ref{tab:benchmark} and visualize key comparisons in Figure~\ref{fig:size_performance}. 
Overall, foundation models consistently outperform from-scratch baselines across both architecture families.
The best foundation model, ECG-FM, achieves a mean AUROC of $0.882$, substantially exceeding the best from-scratch CNN (InceptionTime, $0.812$) and the best from-scratch Transformer (PatchTST, $0.784$).
These results confirm that large-scale pre-training facilitates effective transfer to diverse downstream ECG tasks. 
Despite comparable or even smaller parameter counts, the ResNet-based from-scratch models consistently outperform larger non-ResNet CNNs such as TimesNet ($61.9$M parameters, 0.739) and TCN ($11.7$M parameters, 0.767), suggesting that the ResNet architecture is well suited for ECG representation learning.
Within the ResNet--SL category, we observe a positive correlation between pre-training dataset scale and downstream performance: ECGFounder (10M samples) achieves 0.866, compared to 0.856 for ResNet (Ribeiro) trained on 2.3M samples.
For Transformer--SSL models, performance appears more strongly governed by model size: ECG-FM (90.9M parameters) achieves the highest score of 0.882, while the smaller ViT (Merl) (5.5M parameters) scores 0.805.
Although HeartLang and ViT (Merl) were pre-trained on a similar number of samples, HeartLang is larger in model size than ViT (Merl) and transfers better.
An exception is HuBERT (Base), which, despite its large parameter count (93M) and extensive pre-training data (9.1M samples), achieves only 0.839.
We empirically observe that the BERT-style architecture of HuBERT is sensitive to fine-tuning hyperparameters and tends to underperform relative to its model scale.
Please refer to Appendix~\ref{appendix:scaling_law_for_transfer} for evaluations under the few-shot setting.

\begin{table*}[t]
\caption{Macro AUROC and macro F1 scores on PTB-XL benchmarks. Best and second-best are computed within each architecture group. Fine-tuned models are highlighted in blue.}
\label{tab:ptbxl_results_only}
\centering
% \small
\setlength{\tabcolsep}{4pt} 
\renewcommand{\arraystretch}{1.2}
% \resizebox{\textwidth}{!}{%
\scriptsize
\begin{tabular}{l|cc|cc|cc|cc|cc|c}
\toprule
\multirow{2}{*}{Method} &
\multicolumn{2}{c|}{PTBXL (Diag.)} &
\multicolumn{2}{c|}{PTBXL (Sub)} &
\multicolumn{2}{c|}{PTBXL (Sup)} &
\multicolumn{2}{c|}{PTBXL (Form)} &
\multicolumn{2}{c|}{PTBXL (Rhythm)} &
\multirow{2}{*}{\shortstack{macro-AUC\\(avg)}} \\
\cmidrule(lr){2-11}
& AUC & F1 & AUC & F1 & AUC & F1 & AUC & F1 & AUC & F1 & \\
\midrule
\multicolumn{12}{c}{\textbf{CNN}} \\
XResNet1d101    & 0.9181 & 0.2060 & 0.8982 & 0.3459 & 0.9142 & 0.6997 & 0.8304 & 0.2706 & 0.9588 & 0.3579 & 0.9039 \\
ResNet (Wang)    & \best{0.9322} & \second{0.2396} & \second{0.9290} & \second{0.3653} & \second{0.9266} & \second{0.7304} & 0.8632 & 0.2776 & 0.9450 & 0.3399 & \second{0.9192} \\
InceptionTime   & 0.9264 & 0.2260 & 0.9156 & 0.3395 & 0.9264 & 0.7177 & 0.8419 & 0.2894 & \second{0.9620} & 0.3533 & 0.9145 \\
TimesNet        & 0.8030 & 0.1492 & 0.7825 & 0.2491 & 0.8759 & 0.6554 & 0.6600 & 0.1672 & 0.7958 & 0.1909 & 0.7834 \\
TCN             & 0.8422 & 0.1672 & 0.8441 & 0.2764 & 0.8968 & 0.6819 & 0.6613 & 0.1925 & 0.8567 & 0.2717 & 0.8202 \\
MICN            & 0.7458 & 0.1294 & 0.7858 & 0.2233 & 0.8425 & 0.6176 & 0.6116 & 0.1497 & 0.7859 & 0.2232 & 0.7543 \\
SCINet          & 0.7440 & 0.1179 & 0.7513 & 0.2279 & 0.8399 & 0.6109 & 0.6301 & 0.1572 & 0.7474 & 0.1988 & 0.7425 \\
\cellcolor{ftblue}ResNet (PCLR)
& \ftcell{0.9051} & \ftcell{0.2123} & \ftcell{0.9183} & \ftcell{0.3279} & \ftcell{0.8747} & \ftcell{0.6475} & \ftcell{0.6734} & \ftcell{0.1713} & \ftcell{0.9263} & \ftcell{0.3613} & \ftcell{0.8596} \\
\cellcolor{ftblue}ResNet (Ribeiro)
& \ftcell{0.9030} & \ftcell{0.2060} & \ftcell{0.9027} & \ftcell{0.3399} & \ftcell{0.9166} & \ftcell{0.7053} & \ftcell{0.8533} & \ftcell{0.2686} & \ftcell{0.9543} & \ftcell{\second{0.3755}} & \ftcell{0.9060} \\
\cellcolor{ftblue}ResNet (Merl)
& \ftcell{0.9312} & \ftcell{0.2393} & \ftcell{\best{0.9332}} & \ftcell{\best{0.3752}} & \ftcell{0.9247} & \ftcell{0.7160} & \ftcell{\best{0.8752}} & \ftcell{\second{0.2912}} & \ftcell{0.9144} & \ftcell{0.3232} & \ftcell{0.9157} \\
\cellcolor{ftblue}ECGFounder
& \ftcell{\second{0.9314}} & \ftcell{\best{0.2458}} & \ftcell{0.9264} & \ftcell{0.3651} & \ftcell{\best{0.9334}} & \ftcell{\best{0.7428}} & \ftcell{\second{0.8688}} & \ftcell{\best{0.3301}} & \ftcell{\best{0.9670}} & \ftcell{\best{0.4216}} & \ftcell{\best{0.9254}} \\
\midrule
\multicolumn{12}{c}{\textbf{Transformer}} \\
Transformer    & 0.8282 & 0.1527 & 0.8170 & 0.2628 & 0.8683 & 0.6449 & 0.7037 & 0.1952 & 0.7249 & 0.1521 & 0.7884 \\
Autoformer     & 0.7089 & 0.1203 & 0.7532 & 0.2075 & 0.8098 & 0.5780 & 0.5971 & 0.1368 & 0.6880 & 0.1336 & 0.7114 \\
Informer       & 0.7715 & 0.1273 & 0.7615 & 0.2279 & 0.8478 & 0.6110 & 0.6531 & 0.1599 & 0.6719 & 0.1527 & 0.7412 \\
Reformer       & 0.7736 & 0.1291 & 0.7834 & 0.2331 & 0.8448 & 0.6111 & 0.6899 & 0.1722 & 0.6893 & 0.1489 & 0.7562 \\
FEDformer      & 0.6926 & 0.1334 & 0.7656 & 0.2405 & 0.8424 & 0.6092 & 0.6246 & 0.1524 & 0.7442 & 0.1725 & 0.7339 \\
Crossformer    & 0.8685 & 0.1776 & 0.8769 & 0.2990 & 0.8828 & 0.6643 & 0.7056 & 0.1854 & 0.8313 & 0.2233 & 0.8330 \\
PatchTST       & 0.8263 & 0.1574 & 0.8422 & 0.2748 & 0.8539 & 0.6222 & 0.6650 & 0.1820 & 0.8339 & 0.2196 & 0.8043 \\
iTransformer   & 0.6865 & 0.0985 & 0.7653 & 0.1740 & 0.7686 & 0.5252 & 0.6089 & 0.1541 & 0.7235 & 0.2012 & 0.7106 \\
\cellcolor{ftblue}ViT (Merl)
& \ftcell{\second{0.9078}} & \ftcell{0.2052} & \ftcell{0.9128} & \ftcell{0.3313} & \ftcell{0.9094} & \ftcell{\second{0.6984}} & \ftcell{0.7933} & \ftcell{0.2480} & \ftcell{0.8925} & \ftcell{0.2584} & \ftcell{0.8832} \\
\cellcolor{ftblue}HuBERT (Base)
& \ftcell{0.8927} & \ftcell{0.2040} & \ftcell{0.9054} & \ftcell{0.3275} & \ftcell{0.9040} & \ftcell{0.6780} & \ftcell{0.8139} & \ftcell{0.2577} & \ftcell{\second{0.9448}} & \ftcell{\second{0.3697}} & \ftcell{0.8922} \\
\cellcolor{ftblue}ECG-FM
& \ftcell{0.8962} & \ftcell{\best{0.2192}} & \ftcell{\second{0.9148}} & \ftcell{\best{0.3485}} & \ftcell{\best{0.9163}} & \ftcell{\best{0.7083}} & \ftcell{\best{0.8692}} & \ftcell{\best{0.3030}} & \ftcell{\best{0.9666}} & \ftcell{\best{0.4228}} & \ftcell{\best{0.9126}} \\
\cellcolor{ftblue}HeartLang
& \ftcell{\best{0.9188}} & \ftcell{\second{0.2148}} & \ftcell{\best{0.9181}} & \ftcell{\second{0.3460}} & \ftcell{\second{0.9047}} & \ftcell{0.6885} & \ftcell{\second{0.8582}} & \ftcell{\second{0.2766}} & \ftcell{0.9323} & \ftcell{0.3393} & \ftcell{\second{0.9064}} \\
\bottomrule
\end{tabular}
\end{table*}

\subsection{Fine-tuning ECG foundation models is sample-efficient}
\label{appendix:scaling_law_for_transfer}
\begin{figure}[H]
    \centering
    \includegraphics[width=1.0\linewidth]{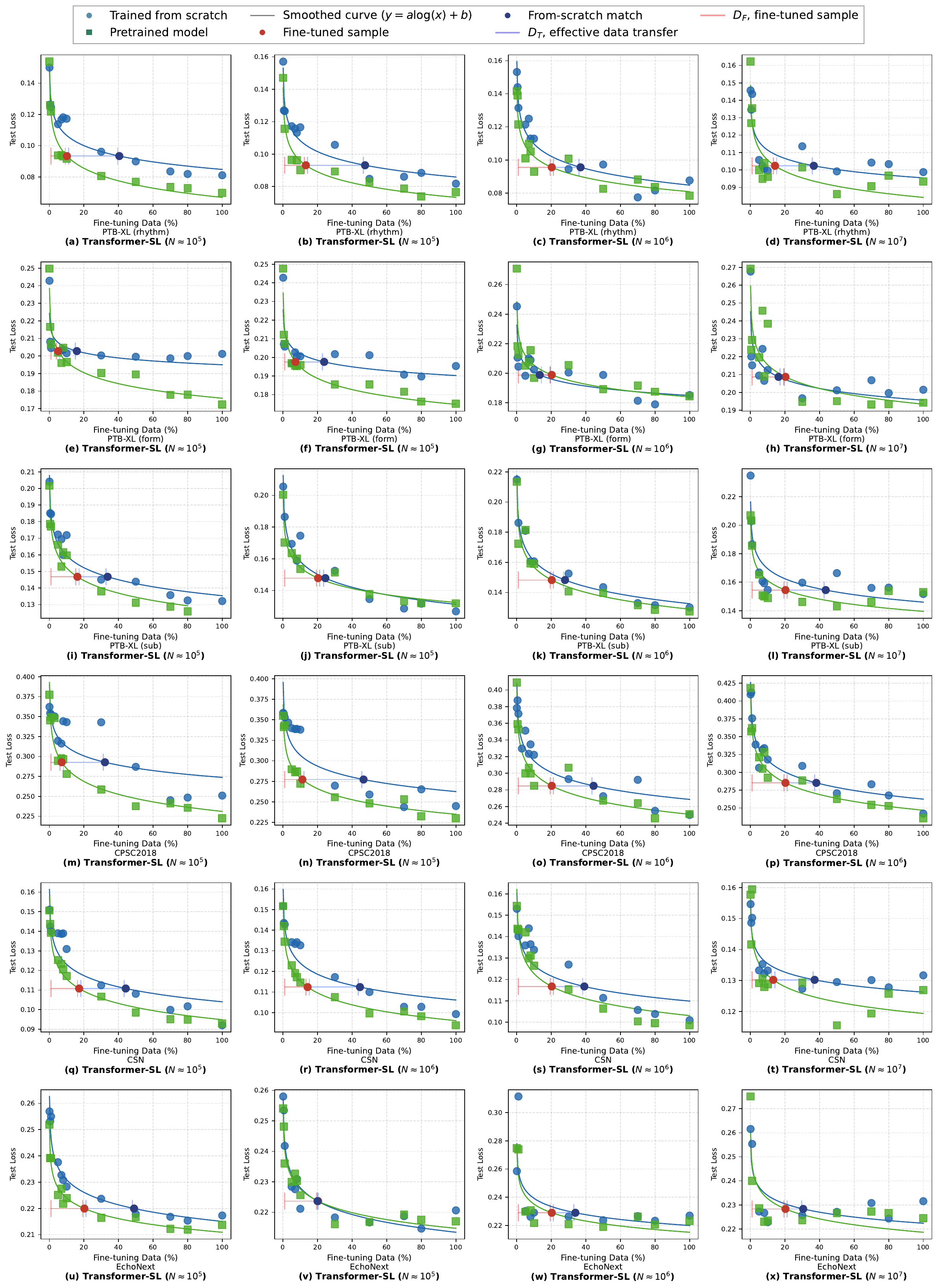}
    \caption{Pretrained Transformer-SL models save downstream data via fine-tuning and facilitate effective data transfer. In each plot, we compare fine-tuning performance against its randomly initialized counterpart.
    We randomly select a fine-tuning ratio to visualize the fine-tuning data used ($D_F$) and the corresponding effective data transfer ($D_T$).}
    \label{fig:transformer_sl_transfer}
\end{figure}

Another question for ECG foundation models is: how much downstream data can we save by fine-tuning a pretrained model, compared to training from scratch?
We study the \emph{scaling law for transfer}~\citep{hernandez2021scaling} on the Transformer-SL model family across six downstream tasks.
We define $D_T$ as the amount of downstream data that fine-tuning effectively ``replaces'', i.e., the gap between the fine-tuning data used ($D_F$) and the from-scratch data needed to reach the same loss.
We model this quantity as a power law in the fine-tuning dataset size and model parameters:
\begin{equation}
    D_T = k \, (D_F)^{\,\gamma_1} \, (N)^{\,\gamma_2},
\end{equation}
where $\gamma_1$ reflects how efficiently additional fine-tuning data translates into transferred knowledge, and $\gamma_2$ captures the effect of model scale on transfer.
We can randomly select multiple observation pairs $(\widehat{D}_F, \widehat{D}_T)$ from the smoothed curve in Figure~\ref{fig:transformer_sl_transfer} and estimate the parameters using ordinary least squares.
We found that fine-tuning consistently reduces the data required to reach a given loss level across all six downstream tasks.
The benefit is most pronounced when the downstream task is closely aligned with the pretraining objective (adult ECG interpretation).
Fitting the power law per dataset, we observe that tasks sharing similar diagnostic standards, CPSC2018 ($\gamma_1{=}1.78$), CSN ($\gamma_1{=}2.59$), and PTB-XL rhythm ($\gamma_1{=}1.85$), exhibit large $\gamma_1$ values, indicating that each additional fine-tuning sample yields substantial equivalent data savings.
In contrast, EchoNext, a structural heart disease dataset further from the pretraining domain, shows a much smaller $\gamma_1{=}0.42$, reflecting limited transferability.
Across all tasks, $\gamma_2$ is consistently negative, suggesting that smaller pretrained models tend to exhibit larger relative data savings.

\section{Limitations and Future Work}
\label{appendix:limitations}
In this work, we pre-train our models on CODE dataset, as it is the second-largest public ECG dataset with six expert-annotated labels. We do not choose the largest one, Harvard-Emory, because its large and machine-generated label sets present a significant challenge for long-tailed multi-label learning in SL models. We also avoid mixing multiple datasets, as different sources (e.g., emergency departments, in-hospital wards, and ambulance settings) typically diverge in data distribution. This divergence poses challenges in unifying preprocessing pipelines and introduces additional inductive bias. The practical challenges of pre-training on such heterogeneous datasets are showcased and discussed in our label scaling experiments.
A limitation of this study is that our conclusions are derived from a single large-scale public dataset rather than a fully heterogeneous multi-source corpus with long-tail distribution.
It is worth noting that while the CODE-II dataset~\citep{abreu2025code} contains a larger set of labels than CODE, it has not yet been publicly released.

Furthermore, this study considers two mainstream ECG models, ResNet and Transformer, as they have proven effective and are widely used as foundation architectures. 
A limitation is that other architectures, such as state-space models, were not systematically evaluated.
Because the specific choice of pre-training paradigm might impact scaling law patterns, we adopt the HeartLang and CMSC methods. 
These approaches are intuitive and simple, similar to vanilla masking and contrastive strategies in natural language processing and computer vision, which we believe introduces minimal inductive bias. Future work includes exploring a more diverse range of masking strategies and sample selection criteria for these pre-training paradigms.

Given that similar conflicting conclusions regarding the non-monotonic scaling behaviors of ID and OOD performance have also emerged in other biosignal data~\citep{yang2026eeg}, the insights from this study hold broader implications for modalities such as electroencephalography (EEG), photoplethysmography (PPG), and electromyography (EMG).

%%%%%%%%%%%%%%%%%%%%%%%%%%%%%%%%%%%%%%%%%%%%%%%%%%%%%%%%%%%%

\end{document}